\documentclass[10pt,twocolumn,letterpaper]{article}

\usepackage{cvpr}
\usepackage{times,microtype}
\usepackage{epsfig}
\usepackage{graphicx}
\usepackage{amsmath}
\usepackage{amssymb}
\usepackage{enumitem}
\usepackage{multirow}
\usepackage{array}
\usepackage{booktabs}
\usepackage{caption,balance}
\usepackage[table]{xcolor}
\usepackage{comment}
\usepackage{xspace,colortbl}
\usepackage{enumitem}
\usepackage{multirow,microtype}
\usepackage{subcaption,microtype}
\usepackage{mathtools}

\newcolumntype{C}[1]{>{\centering\let\newline\\\arraybackslash\hspace{0pt}}m{#1}}
\newcommand{\wraptxt}[1]{#1\xspace}

\newcommand{\tabref}[1]{\wraptxt{Table \ref{#1}}}
\DeclareMathOperator{\argmin}{arg\,min}
\let\oldparagraph\paragraph
\renewcommand{\paragraph}[1]{\oldparagraph{\textnormal{\textbf{#1}}}}

\newcommand{\ds}{\wraptxt{dataset}}
\newcommand{\ml}{\wraptxt{meta-learning}}

\newcommand{\ct}{\wraptxt{catastrophic forgetting}}

\newcommand{\hw}{w_0}

\newcommand{\mi}{\wraptxt{MiniImageNet}}
\newcommand{\ti}{\wraptxt{TieredImageNet}}
\newcommand{\di}{\wraptxt{DomainImageNet}}

\newcommand{\shot}[2]{#1-shot #2-way}

\usepackage[breaklinks=true,bookmarks=false]{hyperref}

\cvprfinalcopy 

\begin{document}





\title{Incremental  Meta-Learning via Indirect Discriminant Alignment}

\author{Qing Liu \thanks{Work conducted while at AWS.}\\
Johns Hopkins University\\
{\tt\small  qingliu@jhu.edu}
\and
Orchid Majumder  \\
AWS\\
{\tt\small orchid@amazon.com}
\and
Alessandro Achille\\
AWS\\
{\tt\small aachille@amazon.com}
\and
Avinash Ravichandran\\
AWS\\
{\tt\small ravinash@amazon.com}
\and
Rahul Bhotika\\
AWS\\
{\tt\small bhotikar@amazon.com}
\and
Stefano Soatto\\
UCLA \& AWS\\
{\tt\small soattos@amazon.com}
}

\maketitle

\newcommand{\C}{\mathcal{C}}
\newcommand{\D}{\mathcal{D}}
\newcommand{\E}{\mathcal{E}}
\newcommand{\T}{\mathcal{T}}
\newcommand{\M}{\mathcal{M}}
\newcommand{\wt}{{w_t}}
\newcommand{\wtp}{{w_{t+1}}}
\newcommand\dw{{\delta w}}

\begin{abstract}
We propose a method to train a model so it can learn new classification tasks while improving with each task solved. This amounts to combining meta-learning with incremental learning. Different tasks can have disjoint classes, so one cannot directly align different classifiers as done in model distillation. On the other hand, simply aligning features shared by all classes does not allow the base model sufficient flexibility to evolve to solve new tasks. We therefore indirectly align features  relative to a minimal set of ``anchor classes.'' Such {\em indirect discriminant alignment} (IDA) adapts a new model to old classes without the need to re-process old data, while leaving  maximum flexibility for the model to adapt to new tasks. This process enables incrementally improving the model by processing multiple learning {\em episodes}, each representing a different learning task, even with few training examples. Experiments on few-shot learning benchmarks show that this incremental approach performs favorably even compared to training the model with the entire dataset at once. 
\end{abstract}

\section{Introduction}
Meta-learning aims to train a model to learn new tasks leveraging knowledge accrued while solving related tasks.  Most  meta-learning methods do not  incorporate experience from learning new tasks to improve the ``base'' (meta-learned) model. Our goal is to enable such improvement, thus creating a virtuous cycle whereby every new task learned enhances the base model in an incremental fashion, without the need to re-process previously seen data. We call this {\em incremental meta-learning} (IML).

While visual classification with a large number of training samples per class has reached performance close to human-level,  learning from few samples (``shots'') remains a challenge, as we discuss in Sect. \ref{sec:discussion}. We explore the hypothesis that incrementally learning a model from a large number of different tasks, each with few training examples, yields a model comparable to one trained with a large number of images at once. Accordingly, we focus on the case of IML for {\em few-shot learning.}

\begin{figure}[t]
    \centering
    \includegraphics[width=1.00\linewidth]{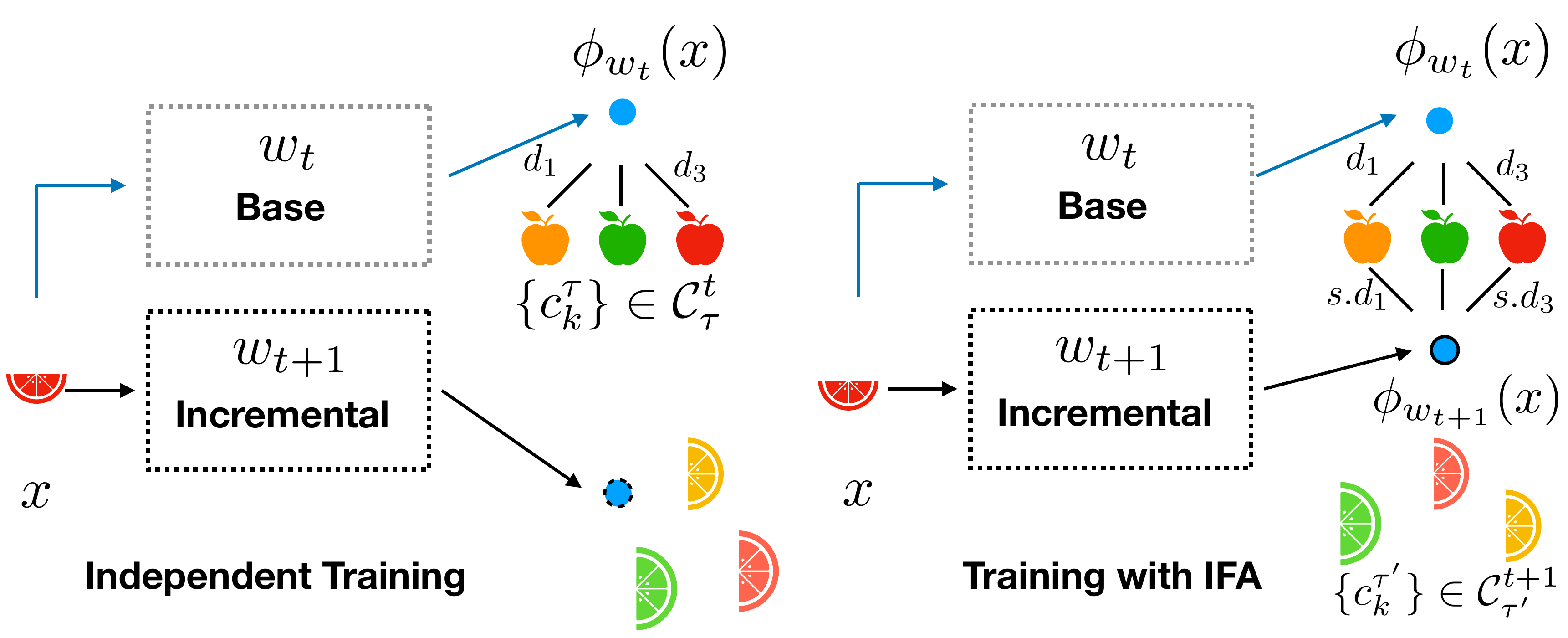}  \hspace{.2cm}
    \caption{\small \sl Indirect Discriminant Alignment (IDA):  Before alignment (left), an orange (new input data) processed through a base model backbone yields an embedding that has a different distance-vector to apples (old class anchors) compared with the one processed through the incremental model backbone. After performing alignment, they produce embeddings that have a similar distance-vector signature, while the incremental model can use the remaining degrees of freedom to adapt the embedding to solve new tasks.}
    \label{fig:illustration}
\end{figure}

IML is not merely  meta-training~\cite{schmidhuber1987evol,bengio1992optimization,snell2017prototypical,finn2017model,ravichandran2019few,lee2019meta} done by processing the training set in chunks: In IML, the meta-training set keeps changing, and we require both the performance of the model for the new tasks, as well as the base model, to improve. IML is also not just incremental (or continual) learning  \cite{rebuffi2017icarl,li2017learning,lopez2017gradient,chaudhry2019efficient,kirkpatrick2017overcoming,schwarz2018progress}, which focuses on a \textit{single}  model to tackle new tasks while avoiding catastrophic forgetting.  In IML, we  want to continuously improve the meta-trained model so that, presented  with an unseen task, it achieves better performance now that it would have before solving the previous task. Moreover, we want to improve the base learner without the need to re-process old data, since that may no longer be accessible, or it may become too expensive to re-process. However, we also want IML to allow exploiting old data, if that is available. Thus far we have used terms like ``training task'' or  ``model'' informally. In the next section, we elaborate on these terms and then make them formal in Sect. \ref{sec:preliminaries}.

\subsection{Nomenclature} 
\label{sec:nomenclature}

We identify a {\em learning task} with a training set. This dataset, together with a function to be minimized ({\em loss}) and the set of functions to minimize it ({\em models}), defines an optimization problem whose solution is {\em a} trained model that ``solves the task.'' So, a learning task can be identified with both a training set, and a suitably trained model. In Sect. \ref{sec:preliminaries}, we will introduce empirical cross-entropy as a loss, deep neural networks (DNNs) as a set of functions, and stochastic gradient descent (SGD) as an optimization scheme. A model consists of a {\em feature representation}, or {\em embedding}, obtained by processing each datum through a {\em backbone}, which is then used to test each hypothesis, or {\em class}, using a {\em discriminant function}.  A  {\em discriminant} is a function that maps a feature to the hypothesis space, where its value is used to render a decision. A discriminant vector is the collection of discriminant values associated to each hypothesis or class. A {\em classifier} is a function that outputs the minimizer (or maximizer) of a discriminant function, which corresponds to a {\em predicted} class or hypothesis. For instance, the Bayesian discriminant is the posterior density of the classes given the data. The corresponding discriminant vector is the collection of posterior probabilities for each hypothesis. Any sufficient statistic of a discriminant is also a discriminant, for instance its negative log or any other invertible function of it.  The optimal Bayesian classifier is one one that returns the hypothesis with the maximum a-posteriori probability (MAP), or with the smallest negative log-posterior.

\subsection{Key Contribution and Organization}

The main contribution of this paper can be more easily understood for the case of metric classifiers, where each class is represented by a prototype, or ``center'' (Fig. \ref{fig:illustration}), and the discriminant compares features to prototypes, for instance using the Euclidean distance in the embedding space, although our method is not restricted to this case. Each new learning task has a set of classes that is possibly disjoint from those of old tasks. The goal of IML is to update the base model incrementally while solving each new task, without necessarily requiring access to data from old tasks, and despite each new task having only few samples per class. Simply imposing that all tasks use the same features would be too restrictive, as different tasks may require the embedding to change while preserving the old centers. On the other hand, we cannot compare discriminants or classifiers directly since they map to different hypothesis spaces. 

Since we cannot compare classifiers directly, and we do not want to needlessly restrict the model's freedom to evolve, the {\bf key idea} is to align models for the old and new tasks {\em indirectly}, by imposing that their discriminants (in the metric case, the vector of distances to the class centers) be aligned {\em relative to a  minimal set of ``anchor classes,''} and otherwise leaving the embedding free to adapt to new tasks. The minimal anchor set is represented by the old centers.  Thus, {\em indirect discriminant alignment} (IDA) is performed by mapping the data to old centers, through both the new and the old backbones, and minimizing the misalignment between the two resulting discriminant vectors. Misalignment can be measured in a number of ways, and the process can be conducted by only processing new data {\em incrementally}, resulting in a continuous improvement of the old model. The more general case, which applies to non-metric classifiers, is explained in  Sect. \ref{sec:IDA}. It results in the {\bf main contribution} of our work, which is to propose what is, to the best of our knowledge, {\em the first method for incremental few-shot meta-learning.}

We tackle the case of few-shot learning since, in the presence of large amounts of data for the classes of interest, pre-training a large model and fine-tuning it for the task at hand already yields a strong baseline. This is not the case for few-shot learning, where the current state-of-the-art still lags far behind \cite{dhillon2019baseline}. Our method directly generalizes several meta-learning algorithms \cite{lee2019meta,snell2017prototypical,ravichandran2019few}, and is applicable to more yet.

In Sect. \ref{sec:implementation} we describe two implementations for performing incremental meta-learning and a number of baselines (Sect.~\ref{sec:baselines}), which form the basis for empirical evaluation in Sect.~\ref{sec:experiments} on few-shot benchmark datasets outlined in Sect.~\ref{sec:datasets}. We also  introduce DomainImageNet in Sect.~\ref{sec:datasets} to measure the effect of meta-learning on incremental learning when new classes are both in- and out-of-domain. We highlight some limitations and failure cases in Sect.~\ref{sec:limitations}. We further discuss related work and future opportunities in Sect.~\ref{sec:discussion}.


\section{Method}
\label{sec:method}

The next section establishes the notation and describes incremental learning and meta-learning in a formalism that will make it easy to describe our key contribution in Sect. \ref{sec:IDA}. At that point, it is a small step to incremental meta-learning, as described in Sect. \ref{sec:incremental}.

\subsection{Preliminaries}
\label{sec:preliminaries}

A model for a classification task is a parametric approximation $p_w(y|x)$ of the posterior distribution of the class $y \in \{c_1, \dots, c_K\}$ given the test datum $x$. For a given class of functions  (architecture), the model may be identified with its parameters ({\em weights}) $w$. The model is trained by minimizing a loss function $L$ that depends on a dataset $\D = \{(x_i, y_i)\}_{i=1}^N$, which defines the task, so that $\hw \doteq \argmin_w L(w; \D)$.

\paragraph{Incremental learning} assumes that an {\em incremental dataset} $\E$ is provided in addition to $\D$. If the two are disjoint, 
we can write 
\begin{equation}
L(w; \D \cup \E) = L(w; \D) + L(w; \E)
\label{eq:disjoint}
\end{equation}
If $L$ is differentiable with respect to $w$ and we train until convergence ($\nabla_w L(\hw, \D) = 0$), we can expand $L$ to second-order around the previous parameters $\hw$ to obtain
\begin{align}
L(w; \D \cup \E) &= L(w; \D) + L(w; \E) \nonumber \\
&\simeq L(\hw + \dw; \E) + L(\hw; \D) \nonumber \\
&+ \dw^T H(\hw; \D) \dw  \nonumber
\end{align}
where $w = \hw + \delta w$ and $H(\hw; \D)$ is the Hessian of the loss $L(w; \D)$ computed at $\hw$. Ignoring the constant term $L(\hw; \D)$ yields the derived loss 
\begin{equation}
\label{eq:linearization}
\mathcal{L}(w) = L(w; \E) + \dw^T H(\hw; \D) \dw
\end{equation}
minimizing which corresponds to fine-tuning the base model for the new task while ensuring that the {\em parameters} change little, using the Hessian as the metric,\footnote{A metric is a positive semi-definite symmetric bilinear form. Since the Hessian $H$ for deep networks typically has some negative eigenvalues, it is typically approximated by the Fisher Information Matrix, which is positive semi-definite and also easier to compute.} a process known as Elastic Weight Consolidation \cite{kirkpatrick2017overcoming}. Note that, even if the incremental set $\E$ is small, there is no guarantee that the weight update $\dw$ will be small. Moreover, making $\dw$ small in eq.~\eqref{eq:linearization} is  unnecessary, since the weights can often change considerably without changing the network behavior.

\paragraph{Distillation} is based on approximating the loss {\em not} by perturbing the weights, $\hw \rightarrow \hw + \dw$, but by perturbing the discriminant function, $p_{\hw} \rightarrow p_{\hw + \dw}$, which can be done by minimizing 
\begin{equation}
    \mathcal{L}(w) =  L(w; \E) + \lambda \mathbb{E}_{x\sim \D} \operatorname{KL}(p_{\hw}(y|x)|| p_w(y|x))
    \label{eq:alignment}
    \end{equation}
where the Kullback-Leibler (KL) divergence measures the perturbation of the new discriminant $p_w$ with respect to the old one $p_{\hw}$ in units $\lambda$. The losses in eq.~\eqref{eq:linearization} and eq.~\eqref{eq:alignment} are equivalent up to first-order, meaning that a local first-order optimization would yield the same initial step when minimizing them. Eq.~\eqref{eq:alignment} may be interpreted as model distillation \cite{ba2014deep,hinton2015distilling} or trust-region optimization  \cite{schulman2015trust}. The drawback of this method is that it needs access to old samples to compute the loss, since the KL is averaged over $\D$. Our goal is to extend these ideas to meta-learning where $\D$ may no longer be accessible.

\paragraph{Meta-Learning} presents an additional difficulty: A meta-training dataset consists of several tasks, indexed by $\tau$, each learned in a separate training {\em episode,} each represented by a different dataset $\D_\tau$ with possibly different classes $\{c_1^\tau, \dots, c_{K}^\tau\} = \C_\tau$. Rather than training a single model to minimize the loss on a single dataset, a meta-learning algorithm aims to produce a task-agnostic model that minimizes the loss across all meta-training tasks. For the case of empirical cross-entropy:
\begin{align}
\label{eq:xent-loss}
L(w; \D) &= \frac{1}{N_\tau} \sum_\tau \frac{1}{|\D_\tau|} \sum_{(x_i, y_i) \in \D_\tau} -\log p_w^\tau(y_i|x_i)
\end{align}
The first sum ranges over however many {\em meta-training tasks} $\D_\tau$ are available. To formalize the core idea in the next section, without loss of generality we write the posterior density $p(y|x)$ in terms of a ``backbone'' function $\phi$ that maps each sample $x$ to a feature vector: $z = \phi(x)$, and a discriminant ``head'' $f$ that maps a feature vector to the posterior $p(y|x) = f(y|\phi(x))$. 

\subsection{Indirect Discriminant Alignment (IDA)} 
\label{sec:IDA}

The challenge in extending incremental learning \eqref{eq:alignment} to meta-learning \eqref{eq:xent-loss} is that each task $\D_\tau$ in the latter has a different discriminant $f_\tau$ for a different set of classes $\C_\tau$. Thus, aligning the discriminants directly would be imposing alignment between different classes, which is not meaningful.

A naive solution would be to just align the features $\phi_w(x)$ on all inputs, for instance by minimizing their average distance
\begin{equation}
\label{eq:DFA}
\mathbb{E}_x\|\phi_{\wt}(x) - \phi_{\wtp}(x)\|^2
\end{equation}
However, this would be needlessly restrictive: Completely different features can yield the same posterior density, and we want to exploit this flexibility for incremental learning. Moreover, we want our method to only process new data, rather than keep re-processing old data, which would defy the goal of incremental processing. To simplify the notation, we refer to $p_{w_\tau}$ as $p_{\rm old}$ and $p_{w_{\tau + 1}}$ as $p_{\rm new}$, and so for the heads $f_{\rm old}, f_{\rm new}$ and the backbones $\phi_{\rm old}, \phi_{\rm new}$. Each can be trained on different tasks, or episodes, $\tau$.

The key idea of this paper is to enable {\em aligning the old and new discriminants using  ``class anchors'' from the old task $\tau$, while processing only data from the new task $\tau'$.} This is done through {\em indirect discriminative alignment} (IDA), illustrated in Fig.~\ref{fig:illustration}, which addresses the  challenge that the tasks $\tau$ and $\tau'$ may not share any classes.
IDA uses the classes defined by the old discriminants as ``anchors,'' and imposes that the features processed through the old and new embeddings share the same discriminant relative to these anchor classes. For metric-based classifiers, the classes can be represented by points in latent (feature, or embedding) space, and the anchors are just an under-complete basis of this space, with the discriminant vector represented by the Euclidean distance to each anchor class representative. The under-complete alignment leaves the residual degrees of freedom free for continual learning. However, the method is more general, allowing any discriminant function.

To make the dependency on the anchor classes and episodes explicit, we write the model $p^\tau_w(y|x)  = f^\tau_w(y|\phi_w(x))$. Indirect discriminative alignment of the new model to the old one is then performed by minimizing:
\begin{equation}
{
    \mathbb{E}_{x \sim \E,\tau'} \big[\operatorname{KL}(f^{\tau'}_{\rm old}(y|\phi_{\rm old}(x) || f^{\tau'}_{\rm old}(y|\phi_{\rm new}(x)))\big] 
}
\end{equation}
where $\C_{\rm old}$ is a set of classes obtained after training on the old training set and  ${\tau'}$ are tasks sampled from the new dataset $\E$.

Intuitively, we reuse the old class representatives $\mathcal{C}_{\rm old}$ and ask that the new features $\phi_{\rm new}$ remain compatible with the discriminant $f_{\rm old}$. Moreover, instead of sampling from the old dataset $\D$ -- which we may no longer have access to -- we sample from $x \sim \E$. In the case of metric classifiers, this can be interpreted as aligning the new features to a set of anchor points, which in particular are the old class representatives.

Note that $f$ can be any discriminant that can process data generated via a representation function $\phi$, where both $f$ and $\phi$ have their own parameters. Also, the choice of KL-divergence to measure the discrepancy between discriminant vectors is due to the fact that it yields a simple expression for most commonly used models, but IDA is not limited to it and any other divergence measure could be employed instead.

\subsection{Incremental Meta-Learning}
\label{sec:incremental}

Given~\eqref{eq:IDA}, incremental meta-learning consists of solving \begin{equation}
    w_{t+1} = \arg\min_{\wtp} L(\wtp; \E) + \lambda \, \operatorname{IDA}_{\E}(\phi_{w_{t+1}} | \phi_{w_{t}}; \C_t )
    \label{eq:IML}
\end{equation}
where the first term  corresponds to fine-tuning the base model on the new data, while the second term enforces indirect discriminant alignment relative to the anchors from old classes. In the next section, we describe our implementation and empirical evaluation.

\subsection{Implementation}
\label{sec:implementation}

The simplest implementation of our method 
 eq.~\eqref{eq:IML} is obtained by using  a metric classifier as the base meta-learner. This choice limits us to each task having the same number of classes, a choice we will discuss and extend in App.~\ref{sec:lifting}. 
We represent a metric-based classifier using a function $\psi_w$ that computes the class representatives, or prototypes, or ``centers,'' $c_k^\tau = \psi_w(\D_\tau)_k$,\footnote{We overload the notation $c$ to indicate the classes in $\C$ and the class representation, which are the argument of $\chi$, since both represent the classes.}
and a function (metric) $\chi_w(z_i, c_k^\tau)$ that scores the fit of a datum, represented by the feature vector $z_i$, with an hypothesis corresponding to a class $c_k^\tau$. 
Each function can be fixed \cite{snell2017prototypical} or learned \cite{ravichandran2019few}.  Note that the backbone $\phi_w$ is common to all tasks, whereas the metric changes with each few-shot task $\D_\tau$, since $c_k^\tau \in \C_\tau$ and $\mathcal{C}_\tau = \{c_k^\tau = \psi_w(\D_\tau)_k\}_{k=1}^K$. According to this model, the optimal (Bayesian) discriminant for the task $\D_\tau$ is of the form:
\begin{align}
    p(y=k|z) = \frac{e^{\chi_w(z, c_k^\tau)}}{\sum_{j} e^{\chi_w(z, c_{j}^\tau)}}
\end{align}
where $z = \phi_w(x)$. Note that $\chi_w$ and $p(y|x)$ are equivalent 
discriminants: Maximizing the posterior is equivalent to minimizing the negative log, which yields a loss of the form 
\begin{align}
\label{eq:xent-loss-expand}
L(w; \D)
    &= \frac{1}{N_\tau} \sum_\tau \frac{1}{|\D_\tau|} \sum_{\mathrlap{(x_i, y_i) \in \D_\tau}}  -\chi_w(z_i, c_{y_i}^\tau) \nonumber \\
    &+ \log \big({\textstyle \sum_{k=1}^K e^{\chi_w(z_i, c_k^\tau)} }\big)
\end{align}
Our first implementation has a trained backbone $\phi_w$ but fixes the metric $\chi$ to be the $L_2$ distance and the class representatives to be the means:
\begin{align*}
\textstyle \chi(z, c) &:= \textstyle - \| z - c \|^2, \\ \textstyle \psi(\D_\tau,k) &:= \textstyle \frac{1}{| \C_k |} \sum_i \delta_{y_i,k} z_i \\\C_\tau &= {\psi(\D_\tau,k)}_{k=1}^K.
\end{align*}
The detailed computation of the loss  eq.~\eqref{eq:xent-loss-expand} is described in the appendix. After every training episode, we discard the data used for meta-training and only retain the class anchors $\C_\tau$. Our paragon (oracle), that will be described in eq.~\eqref{eq:exemplar-alignment}, does not retain any class anchors but trains a new meta-learner at every episode, utilizing all data seen thus far. Ideally, the final performance of the two should be similar, which would justify incremental processing of new datasets without the need to re-process old data, which was our working hypothesis and the basis of eq.~\eqref{eq:linearization}. Indeed, this is what we observe in Sect.  \ref{sec:experiments}.

For meta-training, we sample few-shot tasks using episodic sampling \ie each batch consists of $K$ classes sampled at random. We then sample $N_s$ samples as the support samples and $N_q$ samples as the query samples. The class representations  are calculated only using the support samples, while the query samples are used to compute the loss. For training the base model, we sample few-shot tasks $\tau$ from the old dataset $\D$ and train the model using the loss function in eq.~\eqref{eq:xent-loss}. To train the incremental model we sample a few-shot task $\tau'$ from the new dataset $\E$. We then sample $K$ random class anchors (of the old dataset $\D$) from $C_\tau^{t}$, which are calculated and preserved after the previous training phase. During incremental phase(s), the network is trained by minimizing eq.~\eqref{eq:xent-loss-expand}.

\section{Empirical Validation}

We compare our simplest method, which is based on a Prototypical Network architecture (PN) \cite{snell2017prototypical} as the base meta-learner,  with several baselines as well as the paragon model that uses the same architecture but is free to re-process all past data along with new data. In Sect. \ref{sec:experiments} we assess performance  on standard few-shot image classification benchmarks (\mi and \ti) as well as on a newly curated dataset described in Sect. \ref{sec:datasets}. To show that our method is not tied to the specifics of PN, we also perform the same experiments using ECM \cite{ravichandran2019few}. That is the basis for extending our simplest method to the case where each task has a different number of classes, described in the appendix.

\paragraph{Implementation Details:} 
We use a ResNet-12 \cite{he2016deep} following \cite{oreshkin2018tadam} as our feature extractor $\phi_w$. It consists of four residual layers each with $3\times3$ convolutional layers followed by a max-pooling layer. We use DropBlock regularization \cite{ghiasi2018dropblock}, a form of structured dropout with a keep-rate of $0.9$ after the max-pooling layers. At each round, we train for $200$ epochs, each consisting of $800$ few-shot training tasks containing $5$ ($1$) support examples per class for \shot{5}{5} (\shot{1}{5}). We use 15 query points per class for computing the loss to update the network parameters. Test performance is also measured with 15 query points per class. We use Adam \cite{kingma2014adam} with an initial learning-rate of $0.001$ which is reduced by a factor of $0.5$ when performance on the validation set does not improve for more than $3$ epochs. We use cross-entropy loss with softmax temperature  $2.0$, following \cite{li2017learning}.
For IDA, we choose $\lambda$ to be $1.0$ and we show the effect of varying $\lambda$ in the range of [$0.0$, $10.0$] in the appendix.

\subsection{Baselines and Ablation Studies}
\label{sec:baselines}

To evaluate the method quantitatively, we need an upper-bound (oracle) represented by a model 
that performs meta-training using all the data as well as few other baselines to enable a fair comparison and ablation studies. 

\noindent{\bf No Update (NU)} is the simplest baseline, that is a model meta-trained only using the old dataset.

\noindent{\bf Fine-Tuning (FT)} starts with the model meta-trained on old data and performs additional steps of SGD on the new data wit no additional constraint, using the first term of eq.~\eqref{eq:IML}.

\noindent{\bf Direct Feature Alignment (DFA)} adds to the first term of eq.~\eqref{eq:IML} a penalty for the direct misalignment of features \eqref{eq:DFA} averaged over the new tasks
\begin{equation}
    {\rm DFA}_{\E}(\phi_{w_{t+1}} | \phi_{w_t}) = \mathbb{E}_{x \sim \E{\tau'}} \| \phi_{w_{t+1}(x)} - \phi_{w_t}(x) \|_2^2
    \nonumber
\end{equation}
akin to feature distillation.

\noindent{\bf Exemplar-based incremental meta-learning (EIML)} has access to (possibly a subset of) the old data, so we can add an additional term to eq.~\eqref{eq:IML} to foster tighter alignment via
\small
\begin{align}
   \mathcal{L}(\wtp) &= L(\wtp; \E) \nonumber \\
   &+ \lambda\, \mathbb{E}_{x \in\D_\tau} \big[  \operatorname{KL}(f_{\wt}^\tau(y|\phi_{w_t}(x))|| f_\wtp^\tau(y|\phi_{\wtp}(x)))\big] \nonumber \\
   &+  \lambda\,\mathbb{E}_{x \in\E_{\tau'}} \big[\operatorname{KL}(f_{\wt}^\tau(y|\phi_{\wt}(x))|| f_\wt^\tau(y|\phi_{\wtp}(x)))\big]
   \label{eq:exemplar-alignment}
\end{align}
\normalsize

where $\E_{\tau'}$ is task sampled from the new dataset and $\C_t$ and $\C_{t+1}$ are obtained by re-processing $\D_\tau$ (a task sampled from the old dataset) through the old and the new embeddings respectively. We expect this method to perform best, as it has access to old data. However, it is computationally more expensive than IDA as we need to re-process old data.

\noindent{\bf Full training paragon (PAR)} consists of meta-learning  using the union of  data from old and the new datasets, minimizing the left-hand side of eq.~\eqref{eq:disjoint}. There is no incremental training, so this method serves as an upper bound for performance.

\subsection{Datasets}
\label{sec:datasets}
We test our algorithm on \mi \cite{vinyals2016matching}, \ti \cite{ren2018meta} and another variant of ImageNet \cite{russakovsky2015imagenet} which we call \di. \mi consists of images of size $84 \times 84$ sampled from $100$ classes of the ILSVRC dataset \cite{russakovsky2015imagenet}, with $600$ images per class. We used the data split outlined in \cite{ravi2017optimization}, where $64$ classes are used for training, $16$ classes for validation and $20$ for testing. We further split the $64$ training classes randomly into $32$ for meta-training the base model and the remaining for training the incremental model; $16$ validation classes are only used for assessing generalization during meta-training for both the base and incremental models. For a fair measurement of performance on the old data, we also use a separate test set comprising  300 new images per class \cite{gidaris2018dynamic}.

TieredImageNet is a larger subset of ILSVRC, with $779,165$ images of size $84 \times 84$ representing $608$ classes that are hierarchically grouped into $34$. This dataset is split to ensure that sub-classes within the $34$ groups are not shared among training, validation and test sets. The result is $448,695$ images in $351$ classes for training, $124,261$ images in $97$ classes for validation, and $206,209$ images in $160$ classes for testing. For a fair comparison, we use the same training, validation and testing splits of \cite{ren2018meta} and use the classes at the lowest level of the hierarchy. Similar to \mi, we randomly pick $176$ classes from the training set for meta-training the base model and use the remaining $175$ classes for incremental meta-training. Here we also use a separate test set of about $1000$ images per class for measuring old task performance.

To investigate the role of domain gap in IML, we assemble \di, along the format of \mi, with $32$ old meta-training classes, $32$ new meta-training classes, $16$ meta-validation classes and $40$ meta-test (unseen) classes. All classes are sampled from the ILSVRC dataset, but old, new and meta-test set have two subdivisions, one sampled from \emph{natural} categories, the other sampled from \emph{man-made} categories. $40$ unseen classes consist of $20$ classes each of natural and man-made categories. The domain split we use follows \cite{YosinskiNIPS2104}. 

\subsection{Quantitative Results}
\label{sec:experiments}

We test IML on each \ds using two common few-shot  scenarios: \shot{5}{5} and \shot{1}{5}. We refer to the data used to train the base model as old classes, and that of the incremental model as new classes. We refer to unseen classes as classes that the model has not seen in any training. Final performance of the meta-learner is reported as the mean and $95\%$ confidence interval of the classification accuracy across  $2000$ episodes or few-shot tasks.


\newcommand{\colwidthA}{1.8cm}
\begin{table*}[t!]
\caption{Classification accuracy on 3 different sets: tasks sampled from old, new and unseen classes of \mi using PN \cite{snell2017prototypical} and different IML methods.}
\centering{
\small
\resizebox{0.95\textwidth}{!}{
\begin{tabular}{|l|lll|lll|}
\hline
\multirow{2}{*}{Model} & \multicolumn{3}{c|}{\shot{1}{5}} & \multicolumn{3}{c|}{\shot{5}{5}}  \\
\cline{2-7}
                       & \multicolumn{1}{C{\colwidthA}}{Old classes (32)} & \multicolumn{1}{C{\colwidthA}}{New classes (32)} & \multicolumn{1}{C{\colwidthA}|}{Unseen classes (20)} & \multicolumn{1}{C{\colwidthA}}{Old classes (32)} & \multicolumn{1}{C{\colwidthA}}{New classes (32)} & \multicolumn{1}{C{\colwidthA}|}{Unseen classes (20)}                 \\ 
                       \hline
\rowcolor{orange!20} NU   & $73.84 \pm 0.50$    & $49.05 \pm 0.48$    & $50.55 \pm 0.42$ & $91.17 \pm 0.18$  & $68.35 \pm 0.39$ & $68.60 \pm 0.33$              \\ 
\hline
FT   & $60.64 \pm 0.49$    & $72.61 \pm 0.51$     & $53.60 \pm 0.42$ & $82.25 \pm 0.26$  & $\mathbf{89.63 \pm 0.22}$ & $72.13 \pm 0.33$             \\
DFA      & $60.77 \pm 0.49$    & $72.23 \pm 0.51$     & $53.81 \pm 0.42$ & $82.53 \pm 0.26$  & $89.32 \pm 0.22$ & $72.07 \pm 0.33$                 \\
EIML & $\mathbf{68.95 \pm 0.50}$   & $71.43 \pm 0.52$     & $54.86 \pm 0.42$ & $\mathbf{90.20 \pm 0.20}$  & $86.91 \pm 0.25$ & $74.39 \pm 0.50$            \\
IDA    & $66.54 \pm 0.49$    & $\mathbf{71.92 \pm 0.51}$     & $\mathbf{55.52 \pm 0.43}$ & $89.14 \pm 0.21$  & $87.32 \pm 0.25$ & $\mathbf{75.11 \pm 0.31}$          \\ \hline
\rowcolor{green!20} PAR         & $74.65 \pm 0.49$    & $75.85 \pm 0.50$     & $56.88 \pm 0.43$ & $91.77 \pm 0.17$  & $92.49 \pm 0.17$ & $75.27 \pm 0.13$     \\ \hline
\end{tabular}}}
\label{Tab:miproto}
\end{table*}

\begin{table*}[t!]
\caption{Classification accuracy on 3 different sets: tasks sampled from old, new and unseen classes of \mi using ECM \cite{ravichandran2019few} and different IML   methods.}
\centering{
\small
\resizebox{0.95\textwidth}{!}{
\begin{tabular}{|l|lll|lll|}
\hline
\multirow{2}{*}{Model} & \multicolumn{3}{c|}{\shot{1}{5}} & \multicolumn{3}{c|}{\shot{5}{5}}\\ 
\cline{2-7}
                       & \multicolumn{1}{C{\colwidthA}}{Old classes (32)} & \multicolumn{1}{C{\colwidthA}}{New classes (32) } & \multicolumn{1}{C{\colwidthA}|}{Unseen classes (20)} & \multicolumn{1}{C{\colwidthA}}{Old classes (32)} & \multicolumn{1}{C{\colwidthA}}{New classes (32)} & \multicolumn{1}{C{\colwidthA}|}{Unseen classes (20)}  \\ 
\hline
\rowcolor{orange!20} NU   & $73.82 \pm 0.43$    & $53.00 \pm 0.43$     & $52.77 \pm 0.37$ & $89.38 \pm 0.38$  & $71.90 \pm 0.36$ & $71.57 \pm 0.36$                  \\ \hline
FT    & $63.71 \pm 0.43$    & $75.05 \pm 0.43$     & $56.00 \pm 0.38$ & $82.90 \pm 0.21$  & $89.37 \pm 0.21$ & $74.29 \pm 0.32$                 \\
DFA      & $64.66 \pm 0.42$    & $\mathbf{75.71 \pm 0.43}$     & $56.68 \pm 0.39$ & $83.37 \pm 0.21$  & $\mathbf{89.70 \pm 0.21}$ & $74.69 \pm 0.31$      \\
IDA    & $\mathbf{72.52 \pm 0.42}$    & $68.43 \pm 0.44$     & $\mathbf{57.13 \pm 0.39}$ & $\mathbf{88.46 \pm 0.27}$  & $85.45 \pm 0.27$ & $\mathbf{75.55 \pm 0.30}$            \\\hline
\rowcolor{green!20} PAR        & $74.40 \pm 0.40$    & $75.74 \pm 0.42$     & $59.02 \pm 0.39$ & $89.68 \pm 0.21$  & $89.93 \pm 0.21$ & $77.60 \pm 0.30$          \\ \hline
\end{tabular}}
\label{Tab:miprotopp}
}
\end{table*}

Results of the different methods using PN as a meta-learner are shown in \tabref{Tab:miproto} for \mi, \tabref{Tab:tiproto} for \ti and \tabref{Tab:mim55} for \di. Further, we show results using ECM as a meta-learner in \tabref{Tab:miprotopp} for \mi. We also show the results using ECM on \di for \shot{5}{5} in \tabref{Tab:mim55++}. All results for \di are using natural objects as the old domain and man-made objects as the new domain. In the appendix, we show the results for \shot{1}{5} and also for all combinations of shots and meta-learners while using man-made objects as the old domain and natural objects as the new domain.

\paragraph{Catastrophic Forgetting:} Tables \ref{Tab:miproto},\ref{Tab:miprotopp}, \ref{Tab:tiproto}, \ref{Tab:mim55}, \ref{Tab:mim55++} and \ref{Tab:miprotoincr} show that the classification accuracy on old classes using the incremental model drops significantly when compared with the base model for methods that perform IML without using the old data (\ie FT and DFA). This holds for both \shot{1}{5} and \shot{5}{5}, both PN and ECM, and across all datasets. 

\paragraph{Incremental Meta Learning (IML)} with any of the methods described above yields increased performance on both the new classes and the unseen classes. If performance on the old classes is not a priority,  any IML method will perform better on the new classes with an added bonus of better performance on unseen classes compared with the base model. Again, these conclusions hold across shots, meta-learners and datasets.

\paragraph{EIML vs IDA:} \tabref{Tab:miproto} shows that the difference in  performance of between EIML and IDA is not significant. While we expected EIML to dominate IDA, in some cases EIML performed worse   (\tabref{Tab:miproto}: \shot{1}{5} case for \mi). This illustrates the limited benefit of re-processing old data, justifying IML. We also varied the number of samples we retained from the old dataset in the range of $15$ to $120$ and noticed that the performance was almost constant (shown in the appendix). Hence, we do not run tests on EIML using ECM. Furthermore, for a class of methods that learn the class anchors such as  \cite{lee2019meta}, running EIML is far more expensive as we need to run an additional inner optimization at every step of IML. For completeness, the performance on different datasets using EIML (with PN) is shown in the appendix.

\paragraph{IDA}outperforms all baselines for unseen classes across all scenarios shown in this section,  except for \shot{1}{5} in  \tabref{Tab:tiproto}. We further notice  better performance compared with FT and DFA for old classes. For new classes, IDA trails FT and DFA but overall it performs best on average, approaching the paragon when new tasks are sampled across old, new and unseen classes.

\paragraph{\ti:} \tabref{Tab:tiproto} shows results  using PN \cite{snell2017prototypical} as the meta-learner. For this dataset, the improvement  from using more classes is relatively small compared with \mi (\tabref{Tab:miproto}). When the base model is trained with a large number of classes, the generalization ability of the network is already satisfactory, and we observe negligible \ct or increase in meta-learning performance. We also see that IDA is  similar to the baselines. This raises the question of what new classes would best improve performance in IML. Our experiments on \di address this question. 

\begin{table*}[t!]
\caption{Classification accuracy on 3 different sets: tasks sampled from the old, new and from unseen classes of \ti using PN \cite{snell2017prototypical} and different IML methods.}

\centering{
\small
\resizebox{0.95\textwidth}{!}{
\begin{tabular}{|l|lll|lll|}
\hline
\multirow{2}{*}{Model} & \multicolumn{3}{c|}{\shot{1}{5}} & \multicolumn{3}{c|}{\shot{5}{5}} \\
\cline{2-7}
                       & \multicolumn{1}{C{\colwidthA}}{Old classes (176)} & \multicolumn{1}{C{\colwidthA}}{New classes (175) } & \multicolumn{1}{C{\colwidthA}|}{Unseen classes (160)} & \multicolumn{1}{C{\colwidthA}}{Old classes (176)} & \multicolumn{1}{C{\colwidthA}}{New classes (175) } & \multicolumn{1}{C{\colwidthA}|}{Unseen classes (160)}              \\
                       \hline
\rowcolor{orange!20}  NU   & $73.10 \pm 0.52$    & $66.18 \pm 0.43$     & $56.82 \pm 0.50$ & $89.03 \pm 0.27$  & $81.97 \pm 0.37$ & $75.78 \pm 0.43$                 \\ \hline
FT    & $71.87 \pm 0.52$    & $\mathbf{71.03 \pm 0.52}$     & $58.63 \pm 0.50$ & $87.77 \pm 0.29$  & $\mathbf{87.60 \pm 0.30}$ & $78.20 \pm 0.42$                \\
DFA      & $72.03 \pm 0.51$    & $70.83 \pm 0.53$     & $\mathbf{58.81 \pm 0.50}$ & $87.82 \pm 0.29$  & $87.38 \pm 0.30$ & $78.11 \pm 0.42$                 \\
IDA    & $\mathbf{72.65 \pm 0.51}$    & $70.17 \pm 0.53$     & $58.71 \pm 0.50$ & $\mathbf{89.13 \pm 0.15}$  & $86.91 \pm 0.31$ & $\mathbf{78.40 \pm 0.42}$           \\ \hline
\rowcolor{green!20}  PAR         & $78.57 \pm 0.51$    & $77.43 \pm 0.50$     & $61.87 \pm 0.51$ & $91.05 \pm 0.24$  & $90.44 \pm 0.26$ & $80.58 \pm 0.40$          \\ \hline
\end{tabular}}
\label{Tab:tiproto}
}
\end{table*}

\paragraph{DomainImageNet:} Results for \shot{5}{5} are shown for PN \cite{snell2017prototypical}  in \tabref{Tab:mim55} and for ECM \cite{ravichandran2019few}  in \tabref{Tab:mim55++}. The model is first trained using natural classes and then  incrementally trained using man-made classes. This helps evaluate the effect of domain shift between old and new training classes. We test on five different sets: seen and unseen classes from natural objects, seen and unseen classes from man-made objects and unseen classes from a mixture of the two. 

The tables show that the accuracy on the joint test set improves significantly compared with the baselines. Most of the gain is for the new domain, \ie, man-made objects. Also, \ct is significant since there is domain shift between the classes from the old and new domains. This effect is also seen   with unseen classes on the same domain. IDA is shows improvement across the board relative to the baselines. The results for \shot{1}{5} and using the reverse domain training (\ie, old domain is man-made objects and incremental domain is natural objects) on the three sets for all IML algorithms show similar trends. This suggests that it matters what classes are selected for incremental training. Adding classes with diverse statistics yields maximum advantage.  While we can expect this to be the trend for samples belonging to the same class, we find it to be true for samples belonging to unseen classes as well from the same domain. Our method successfully mitigates \ct to a large extent and performs well across different domains.

\begin{table*}[t!]
\caption{Results of 5-shot 5-way classification accuracy on different sets of \di using PN \cite{snell2017prototypical} and different IML methods.}
\centering{
\small
\resizebox{0.85\textwidth}{!}{
\begin{tabular}{|l|ll|ll|l|}
\hline
                      \multicolumn{1}{|C{\colwidthA}|}{Model} & \multicolumn{1}{C{\colwidthA}}{Old classes from old domain (32)} & \multicolumn{1}{C{\colwidthA}|}{New classes from new domain (32)} & \multicolumn{1}{C{\colwidthA}}{Unseen classes from old domain (20)} & \multicolumn{1}{C{\colwidthA}|}{Unseen classes from new domain (20)} & \multicolumn{1}{C{\colwidthA}|}{Unseen classes from both domains (40)}                \\
                       \hline
\rowcolor{orange!20} NU   & $86.94 \pm 0.22$    & $49.14 \pm 0.36$     & $57.66 \pm 0.38$     & $51.72 \pm 0.32$     & $59.59 \pm 0.35$ \\ \hline
FT    & $64.42 \pm 0.35$    & $\mathbf{84.80 \pm 0.28}$     & $50.72 \pm 0.38$     & $71.16 \pm 0.32$     & $65.44 \pm 0.40$ \\
DFA      & $65.12 \pm 0.35$    & $83.95 \pm 0.29$     & $51.33 \pm 0.38$     & $70.46 \pm 0.33$     & $65.52 \pm 0.40$ \\
IDA    & $\mathbf{81.26 \pm 0.27}$    & $82.06 \pm 0.30$     &  $\mathbf{59.32 \pm 0.39}$     & $\mathbf{70.61 \pm 0.32}$     & $\mathbf{70.36 \pm 0.36}$ \\ \hline
\rowcolor{green!20} PAR & $87.44 \pm 0.22$    & $88.77 \pm 0.25$     & $58.59 \pm 0.37$     & $74.46 \pm 0.32$     & $74.02 \pm 0.37$ \\ \hline
\end{tabular}}
\label{Tab:mim55}
}
\end{table*}

\begin{table*}[t!]
\caption{Results of 5-shot 5-way classification accuracy on different sets of \di using ECM \cite{ravichandran2019few} with different IML methods.}
\centering{
\small
\resizebox{0.85\textwidth}{!}{
\begin{tabular}{|l|ll|ll|l|}
\hline
                      \multicolumn{1}{|C{\colwidthA}|}{Model} & \multicolumn{1}{C{\colwidthA}}{Old classes from old domain (32)} & \multicolumn{1}{C{\colwidthA}|}{New classes from new domain (32)} & \multicolumn{1}{C{\colwidthA}}{Unseen classes from old domain (20)} & \multicolumn{1}{C{\colwidthA}|}{Unseen classes from new domain (20)} & \multicolumn{1}{C{\colwidthA}|}{Unseen classes from both domains (40)}                \\
                       \hline
\rowcolor{orange!20} NU   & $87.86 \pm 0.20$    & $56.71 \pm 0.39$     & $63.30 \pm 0.38$     & $58.10 \pm 0.35$     & $66.09 \pm 0.35$  \\ \hline
FT     & $67.35 \pm 0.34$    & $\mathbf{89.68 \pm 0.20}$     & $55.37 \pm 0.38$     & $\mathbf{74.00 \pm 0.31}$     & $69.98 \pm 0.39$  \\
DFA      & $69.33 \pm 0.33$    & $88.72 \pm 0.22$     & $57.06 \pm 0.38$     & $73.97 \pm 0.31$     & $70.77 \pm 0.38$  \\
IDA    & $\mathbf{86.09 \pm 0.22}$    & $81.82 \pm 0.28$     & $\mathbf{64.22 \pm 0.38}$     & $69.92 \pm 0.33$     & $\mathbf{72.64 \pm 0.33}$  \\ \hline
\rowcolor{green!20} PAR & $86.83 \pm 0.22$    & $88.84 \pm 0.21$     & $65.77 \pm 0.38$     & $75.98 \pm 0.31$     & $77.31 \pm 0.33$  \\ \hline
\end{tabular}}

\label{Tab:mim55++}
}
\end{table*}

\paragraph{Multiple Rounds of IML:} In the above experiments, our configuration consists of one old and one new dataset. In Table \ref{Tab:miprotoincr}, we show the performance of different IML algorithms for a scenario where there are multiple new datasets. We split the new classes of \mi into two sets each having 16 classes (classes are split randomly) and run IML for a $5$-shot $5$-way setup using PN. From the table, we can observe that IDA does not incur any performance loss and achieves similar accuracy on the unseen classes compared to a single training with 32 classes. For other methods like DFA and FT, we can observe some performance drop when comparing with Table \ref{Tab:miproto}, which shows that IDA scales better beyond single incremental training.

\begin{table*}[t!]
\caption{Results of 5-shot 5-way classification accuracy on \mi using PN \cite{snell2017prototypical} with $2$ rounds of incremental meta-training, where each round consists of an $16$ new classes.}
\centering{
\footnotesize
\resizebox{0.95\textwidth}{!}{
\begin{tabular}{|l|lll|lll|}
\hline
\multirow{2}{*}{Model} & \multicolumn{3}{c|}{Incremental - Round I} & \multicolumn{3}{c|}{Incremental - Round II} \\
\cline{2-7} &
                      \multicolumn{1}{C{\colwidthA}}{Old classes ($32$)} & \multicolumn{1}{C{\colwidthA}}{New classes ($16$)} & \multicolumn{1}{C{\colwidthA}|}{Unseen classes ($20$)} & \multicolumn{1}{C{\colwidthA}}{Old classes ($32$+$16$)} & \multicolumn{1}{C{\colwidthA}}{New classes ($16$)} & \multicolumn{1}{C{\colwidthA}|}{Unseen classes ($20$)} \\
\hline
\rowcolor{orange!20} NU  & $91.17 \pm 0.18$    & $65.60 \pm 0.39$     & $68.60 \pm 0.33$ & $82.25 \pm 0.37$  & $71.45 \pm 0.38$ & $68.60 \pm 0.33$                 \\ 
\hline
FT    & $80.70 \pm 0.31$    & $87.67 \pm 0.37$     & $67.45 \pm 0.37$ & $76.03 \pm 0.36$  & $90.72 \pm 0.23$ & $70.57 \pm 0.32$              \\
DFA      & $\mathbf{87.69 \pm 0.26}$    & $88.43 \pm 0.36$     & $68.20 \pm 0.36$ & $80.69 \pm 0.38$  & $91.27 \pm 0.21$ & $71.19 \pm 0.37$         \\
IDA    & $87.30 \pm 0.25$    & $\mathbf{89.56 \pm 0.20}$     & $\mathbf{72.08 \pm 0.36}$ & $\mathbf{84.21 \pm 0.30}$  & $\mathbf{93.25 \pm 0.17}$ & $\mathbf{75.15 \pm 0.35}$   \\
\hline
\rowcolor{green!20} PAR  & $93.94 \pm 0.05$    & $93.09 \pm 0.06$     & $72.10 \pm 0.13$ & $93.03 \pm 0.06$  & $95.58 \pm 0.05$ & $75.27 \pm 0.13$          \\ \hline
\end{tabular}}
\label{Tab:miprotoincr}
}
\end{table*}

\subsection{Limitations and Failure Cases}
\label{sec:limitations}
The implementation we chose, based on \cite{snell2017prototypical} and \cite{ravichandran2019few}, and the tests we performed limit our assessment to tasks that share the same number of classes, $K = 5$, as customary in the literature. While technically not a limitation as one could always build a set of models, each for a different number of classes, and indeed it is not uncommon to train and fine-tune different models for different ``ways'' as seen in the literature, we use the same model for all tests. It is nonetheless desirable  to have a meta-learner that can handle an arbitrary number of classes, different for each training episode. While our general framework eq.~\eqref{eq:IML} enables it, our simplest implementation described in eq.~\eqref{sec:implementation} does not. In appendix, section \ref{sec:lifting}, however, we describe a modified implementation that is not subject to this restriction.  Since benchmarks in the literature most commonly refer to the cases $K=1,5$, we use the simpler model in our experiments. 

Further, sampling $K$ classes among many has low probability of yielding hard tasks that can be informative of meta-learning. Even simple classifiers can easily tell 5 random classes from ImageNet apart. {\em Hard task mining} could be done by selecting tasks using a distance such as  \cite{achille2019task2vec},  by sampling a random class and picking the $4$ closest ones in Task2Vec space for a $5$-way setup.

Finally, in our experiments we have noticed that there is still a performance gap between IDA and the paragon. The performance is matched for the case of unseen classes, but there is room for improvement in tasks sampled from new/current task distribution across shots, datasets and methods.

\section{Discussion and Related Work}
\label{sec:discussion}
The natural occurrence of classes in the world is a long tailed distribution \cite{Zhu:CVPR14,van2017devil,wangNips17}, whereby instances for most classes are rare and instances for few classes are abundant. Deep neural networks trained for classification \cite{Krizhevsky:2012,he2016deep,Huang_2017_CVPR} do not fare well when trained with small datasets typical of the tail \cite{van2017devil}, leading to increased interest in few-shot learning. Meta-learning \cite{thrun2012learning,naik1992meta,schmidhuber1987evol} for few-shot learning \cite{ravi2017optimization,finn2017model,rusu2018meta,vinyals2016matching,snell2017prototypical,lee2019meta,ravichandran2019few,gidaris2018dynamic} uses a relatively ``meta training'' dataset from which several few-shot tasks are sampled to mimic phenomena at the tail. Once meta-trained on the old dataset, these methods cannot take advantage of the new few-shot tasks to update the meta-learner. The obvious fix, to re-train the meta-learner every time a few-shot task arises, is impractical if at all possible, as one may not have access to all past data. 

Incremental learning, or continual learning, is typically performed by adapting a neural network model, trained using some dataset, using a new dataset, to arrive at a single model. The main challenge here is to prevent catastrophic forgetting~\cite{french1999catastrophic}. A few relevant works in this area include \cite{rebuffi2017icarl,li2017learning,lopez2017gradient,chaudhry2019efficient,kirkpatrick2017overcoming,schwarz2018progress}. To learn classifiers in a class-incremental way where new classes are added progressively, \cite{rebuffi2017icarl} proposed to keep exemplars from old classes based on their representation power and the computational budget. \cite{li2017learning} used knowledge distillation to preserve the model's capabilities on the old tasks when only new data is accessible. \cite{lopez2017gradient} and its extension \cite{chaudhry2019efficient} leveraged a small episodic memory to alleviate forgetting. \cite{kirkpatrick2017overcoming} slowed down learning on the weights that were important for old tasks, while \cite{schwarz2018progress} extended it by training a knowledge-base in alternating phases. 

Methods that used few-shot incremental sets such as \cite{gidaris2018dynamic} unified the recognition of both new and old classes using attention based few-shot classification. Similarly, \cite{ren2018incremental} used recurrent back-propagation to train a set of new weights to achieve good overall classification on both old and new classes. \cite{xiang2019incremental,siam2019adaptive} extended this class-incremental framework to other visual recognition tasks like semantic segmentation and attribute recognition. Accordingly, despite being called incremental few-shot learning, these methods are more accurately described as incremental learning using few-shot datasets. 

On-line meta-learning \cite{finn2019online}  can be done by exposing an agent to new tasks in a sequential manner. One may see this experimental setup to be similar to ours; however, unlike IML, \cite{finn2019online} retains data from all previous tasks and leverages it for meta-training, thus forgoing incremental learning. In our  experimental setup, we retain  minimal amounts of data from the old training set. \cite{jerfel2019reconciling,harrison2019continuous} on the other hand, tackled a continual/online meta-learning setup where an explicit delineation between different tasks is not available, whereas in our experimental setup we are primarily trying to solve new classification tasks with clear task boundaries.

To summarize, there are several approaches to solve IML: One that biases new weights to remain similar to those of the base model (elastic weight consolidation) as in eq.~\eqref{eq:linearization},  and one that looks at function space and imposes that the activations remain similar to that of the base model (knowledge distillation), as in eq.~\eqref{eq:xent-loss}.  We adopt the latter and  empirically test how our general framework performs in the case of two metric-based meta-learners \cite{snell2017prototypical,ravichandran2019few}. This is a particular instance of the right-hand side of eq.~\eqref{eq:disjoint}, that was our starting point in Sect. \ref{sec:preliminaries}. It yields empirical performance comparable to meta-learning on the union of the old and new datasets, which is the gold standard. This gives empirical validation to our method, and to the many possible variants that can be explored considering combinations of meta and few-shot set, choices of metrics, classifiers, divergence measures, and a myriad of other ingredients in the IML recipe to minimize eq.~\eqref{eq:IML}. We have tested several options in our experiments and in the appendix, and many more are open for investigation in future work.

{\small
\balance
\bibliographystyle{ieee_fullname}
\bibliography{egbib}
}

\appendix
\clearpage
\onecolumn

\section*{\hfil Appendix \hfil}

In this appendix we report empirical comparison between Indirect Discriminant Alignment (IDA) and Exemplar Incremental Meta-Learning (EIML) for all three datasets - MiniImageNet, TieredImageNet and DomainImageNet with PN as the meta-learner. We also report performance of IDA on different tasks by varying $\lambda$. Along the same lines, we show the performance of EIML as we vary the number of exemplars stored from the old task distribution. Additional empirical results on \di dataset to better portray the impact of the domain gap. Performance comparison of different \ml algorithms when the ``shots'' and ``ways'' vary between meta-training and few-shot testing are also shown.

\section{Full expression for the loss function \label{sec:Derivation}}

The full expression for the loss function defined in eq.~\eqref{eq:xent-loss-expand} for Prototypical Networks (PN) \cite{snell2017prototypical} is obtained as follows: Define
\begin{align*}
\textstyle \chi(z, c) &:= \textstyle - \| z - c \|^2 \\ \textstyle \psi(\D_\tau,k) &:= \textstyle \frac{1}{| \C_k |} \sum_i \delta_{y_i,k} z_i \\\C_\tau &= {\psi(\D_\tau,k)}_{k=1}^K
\end{align*}
Using these choices, we can rewrite $f_w(y=k|x, \C)$  as 
\begin{align}
    f_w(y=k|x, \C) = \frac{e^{- \| z - c_k \|^2}}{\sum_{j} e^{- \| z - c_j \|^2}}
\end{align}
Applying the negative logarithm on both sides we get
\begin{align}
    -\log (f_w(y=k|x, \C)) &=  \| z - c_k \|^2 + \log {\sum_{j} e^{- \| z - c_j \|^2}} \nonumber \\
    & = \| z - c_k \|^2 + \operatorname{LSE}(z,\C)
\end{align}
where $\operatorname{LSE}(z,\C) =\log {\sum_{c\in \C} e^{- \| z - c \|^2}}$.
The new model weights $\wtp$ can then be obtained by solving the following optimization problem
\begin{align}
    \arg\min_{\wtp} \mathcal{L}(\wtp) =  &\arg\min_{\wtp}  L(\wtp; \E) 
    + \lambda \, \operatorname{IDA}_{\E}(\phi_{w_{t+1}} | \phi_{w_{t}}; \C^t )
\end{align}
where the Indirect Feature Alignment (IFA) loss is defined by
\begin{equation}
    \operatorname{IFA}_{\E} (\phi_{\wtp} |  \phi_\wt;  \C^t)  = \mathbb{E}_{x \sim \E{\tau'},\hat \C \sim \mathcal{C}^t} \big[  \operatorname{KL}(f_\wtp(y|x, \hat \C))|| f_\wt(y|x, \hat \C)\big]
    \label{eq:IDA}
\end{equation}
where
\begin{align}
    f_\wtp(y=k|x, C_{\tau}^{t}) &= \frac{e^{- \| \phi_\wtp(x) - c_k^t \|^2}}{\sum_{j} e^{- \| \phi_\wtp(x) - c_j^t \|^2}} \label{eq:newmodel}\\
    f_\wt(y=k|x, C_{\tau}^{t}) &= \frac{e^{- \| \phi_\wt(x) - c_k^t \|^2}}{\sum_{j} e^{- \| \phi_\wt(x) - c_j^t \|^2}}
    \label{eq:oldmodel}
\end{align}
Eq \eqref{eq:oldmodel} shows the discriminant calculated using the old  embeddings, the old class centers and the input as new classes. Similarly, \eqref{eq:newmodel} shows the discriminant calculated using the new model embeddings, the old class centers and the input as  new classes.

The cross-entropy loss $L(w; \E)$ can be rewritten explicitly in the case of prototypical networks as:
\begin{align}
L(w; \E) &= \frac{1}{N_\tau} \sum_\tau \frac{1}{|\E_\tau|} \sum_{(x_i, y_i) \in \E_\tau} -\log p_w^\tau(y_i|x_i)\\
    &= \frac{1}{N_\tau} \sum_\tau \frac{1}{|\E_\tau|} \sum_{\mathrlap{(x_i, y_i) \in \E_\tau}}  \|z_i -c_{y_i}^\tau\|^2 + \operatorname{LSE} (z_i,\C)
\end{align}
We use this as our loss when we meta-train with PN. The loss for ECM is identical, with the caveat that the class identities $c_\tau^t$ are not sample means but are instead learned via optimization \cite{ravichandran2019few}.

\section{Comparison of Exemplar Incremental Meta-Learning and Indirect Discriminant Alignment}
\label{sec:EIMLvsIDA}
In  \tabref{Tab:comparison}, we show the performance comparison between EIML and IDA on all datasets we considered --- both in the  \shot{5}{5} and \shot{1}{5} setup --- using prototypical network as the meta-learner. While we expected EIML to perform better than IDA on all scenarios due to availability of additional samples from the old task distribution, we actually observed that IDA outperforms EIML on unseen classes and performs equally well on few-shot tasks containing new classes. However, as expected, EIML performs better when it comes to handling tasks from the old task distribution.

\renewcommand{\colwidthA}{1.8cm}
\begin{table*}[t!]
\caption{Comparison of EIML and IDA across multiple datasets}
\centering{
\resizebox{0.95\textwidth}{!}{
\begin{tabular}{l|l|lll|lll}
\hline
\multirow{2}{*}[-0.5em]{\ds} &  \multirow{2}{*}{Method} & \multicolumn{3}{c|}{\shot{1}{5}} & \multicolumn{3}{c}{\shot{5}{5}}   \\
\cline{3-8}
                       & & \multicolumn{1}{C{\colwidthA}}{Old classes } & \multicolumn{1}{C{\colwidthA}}{New classes  } & \multicolumn{1}{C{\colwidthA}|}{Unseen classes } & \multicolumn{1}{C{\colwidthA}}{Old classes } & \multicolumn{1}{C{\colwidthA}}{New classes } & \multicolumn{1}{C{\colwidthA}}{Unseen classes }\\ 
                       \hline 
\multirow{2}{*}{\mi} 
& EIML & $\mathbf{68.95 \pm 0.50}$   & $71.43 \pm 0.52$     & $54.86 \pm 0.42$ & $\mathbf{90.20 \pm 0.20}$  & $86.91 \pm 0.25$ & $74.39 \pm 0.50$ \\
& IDA    & $66.54 \pm 0.49$    & $\mathbf{71.92 \pm 0.51}$     & $\mathbf{55.22 \pm 0.43}$ & $89.14 \pm 0.21$  & $\mathbf{87.32 \pm 0.25}$ & $\mathbf{75.11 \pm 0.31}$ \\
\hline
\multirow{2}{*}{\ti}  
& EIML & $72.50 \pm 0.51$    & $69.44 \pm 0.52$     & $58.42 \pm 0.50$ & $88.93 \pm 0.27$ & $86.25 \pm 0.32$ & $77.97 \pm 0.42$ \\
& IDA    & $\mathbf{72.65 \pm 0.51}$    & $\mathbf{70.17 \pm 0.53}$     & $\mathbf{58.71 \pm 0.50}$ & $\mathbf{89.03 \pm 0.27}$  & $\mathbf{86.91 \pm 0.31}$ & $\mathbf{78.40 \pm 0.42}$     \\ 
\hline
\multirow{2}{*}{\di} 
& EIML & $\mathbf{48.48 \pm 0.44}$ & $\mathbf{42.27 \pm 0.41}$ &$ 43.91 \pm 0.42$ &  $\mathbf{83.23 \pm 0.25}$ & $\mathbf{65.81 \pm 0.40}$ & $69.99\pm 0.36$  \\
& IDA & $42.57 \pm 0.41$ & $38.95 \pm 0.39$ & $\mathbf{44.88 \pm 0.43}$ & $81.26 \pm 0.27$ & $65.78 \pm 0.4$1 & $\mathbf{70.36\pm 0.36}$\\
\hline
\end{tabular}}
\label{Tab:comparison}
}
\end{table*}

\section{Varying $\lambda$ in Indirect Discriminant Alignment}
\begin{figure}[h]
\centering
\includegraphics[width=0.6\linewidth]{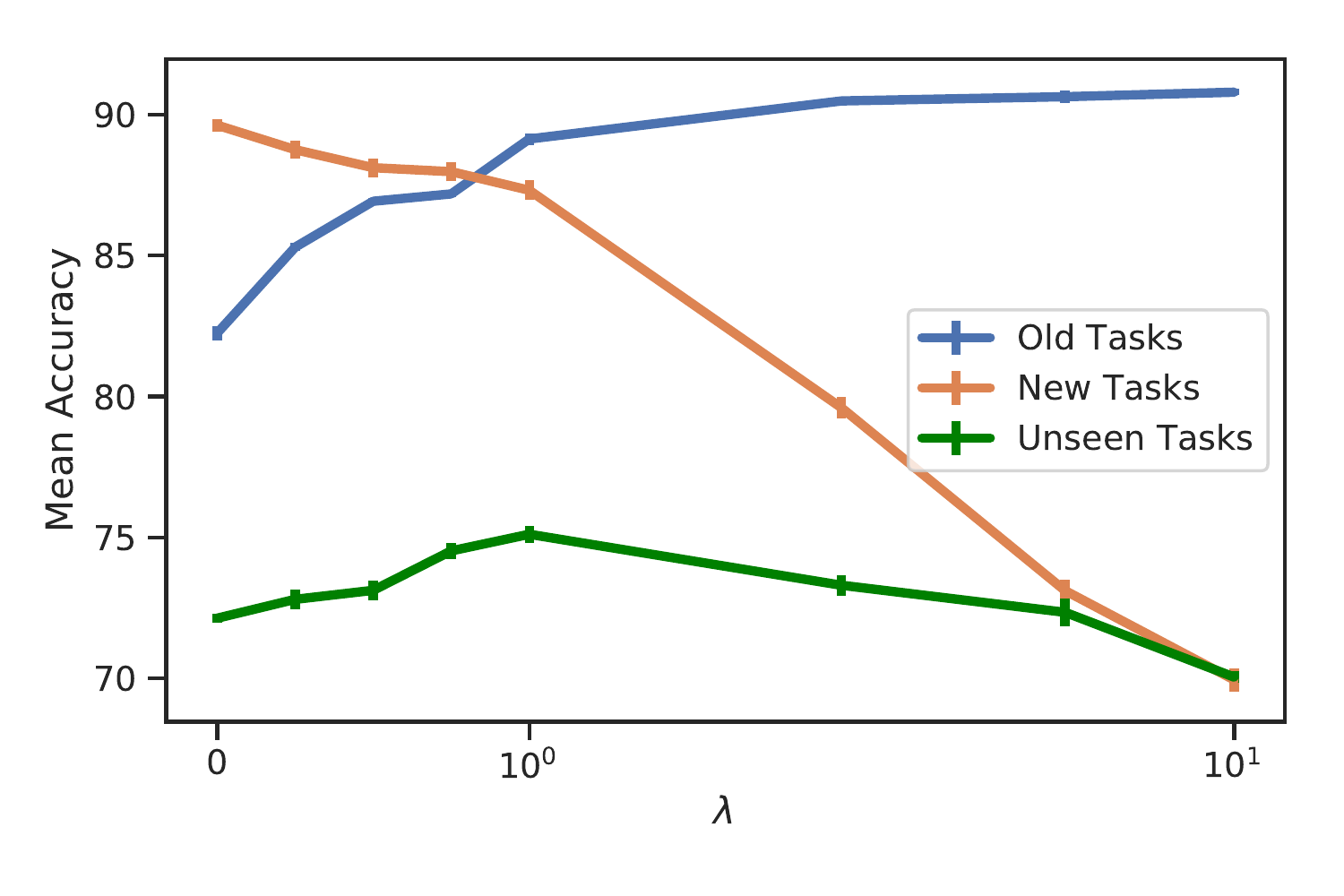}
\caption{Performance of the model on different few-shot tasks: old, new and unseen with $\lambda$ varying from $0.0$ to $10.0$. The figure shows the mean accuracy averaged over 2000 episodes. The errorbars indicate the $95\%$ confidence interval.}
\label{Fig:lambda_percentage}
\end{figure}

By varying $\lambda$ for IDA (Eq. 9), we can maintain a trade-off between the IDA loss and the standard meta-learning loss. In this experiment, we investigate how changing the value of $\lambda$ affects the model's performance on old, new and the unseen test set. We investigated this in the \shot{5}{5} setup for MiniImageNet using PN. We chose $\lambda$ $\in$ $\{0.0, 0.25, 0.5, 0.75, 1.0, 2.0, 5.0, 10.0\}$. All numbers reported in the main paper were with $\lambda=1.0$. The outcome of this experiment can be visualized in Fig. \ref{Fig:lambda_percentage} where we can see that as $\lambda$ is increased, the model's performance on old tasks improves while its performance deteriorates on the new set of tasks. This is intuitive as IDA loss adds a constraint for the incremental model to be similar to the model learned from old tasks and increasing its contribution in the overall loss enhances the model's ability to perform better on old tasks. However, for the same reason, it diminishes the impact of the standard meta-learning loss and hampers performance on new tasks. At $\lambda=0.0$, the model only trains with the standard meta-learning loss and its performance is identical to our baseline named ``Fine-Tuning (FT)''. At $\lambda$ value of $10.0$, IDA loss dominates and the performance is similar (not fully identical though) to our ``No Update (NU)'' baseline which is the model trained only using old tasks. While measuring performance on the unseen  clases, we observe that the ideal value of $\lambda$ lies somewhere between the values of $\lambda$ that provide the best performance on the old tasks and the values of $\lambda$ that work best for the new tasks. The farther $\lambda$ strays from this range, the more it degrades on the unseen tasks.

\section{Varying number of samples in Exemplar Incremental Meta-Learning}
\begin{figure}[h]
\centering
\includegraphics[width=0.6\linewidth]{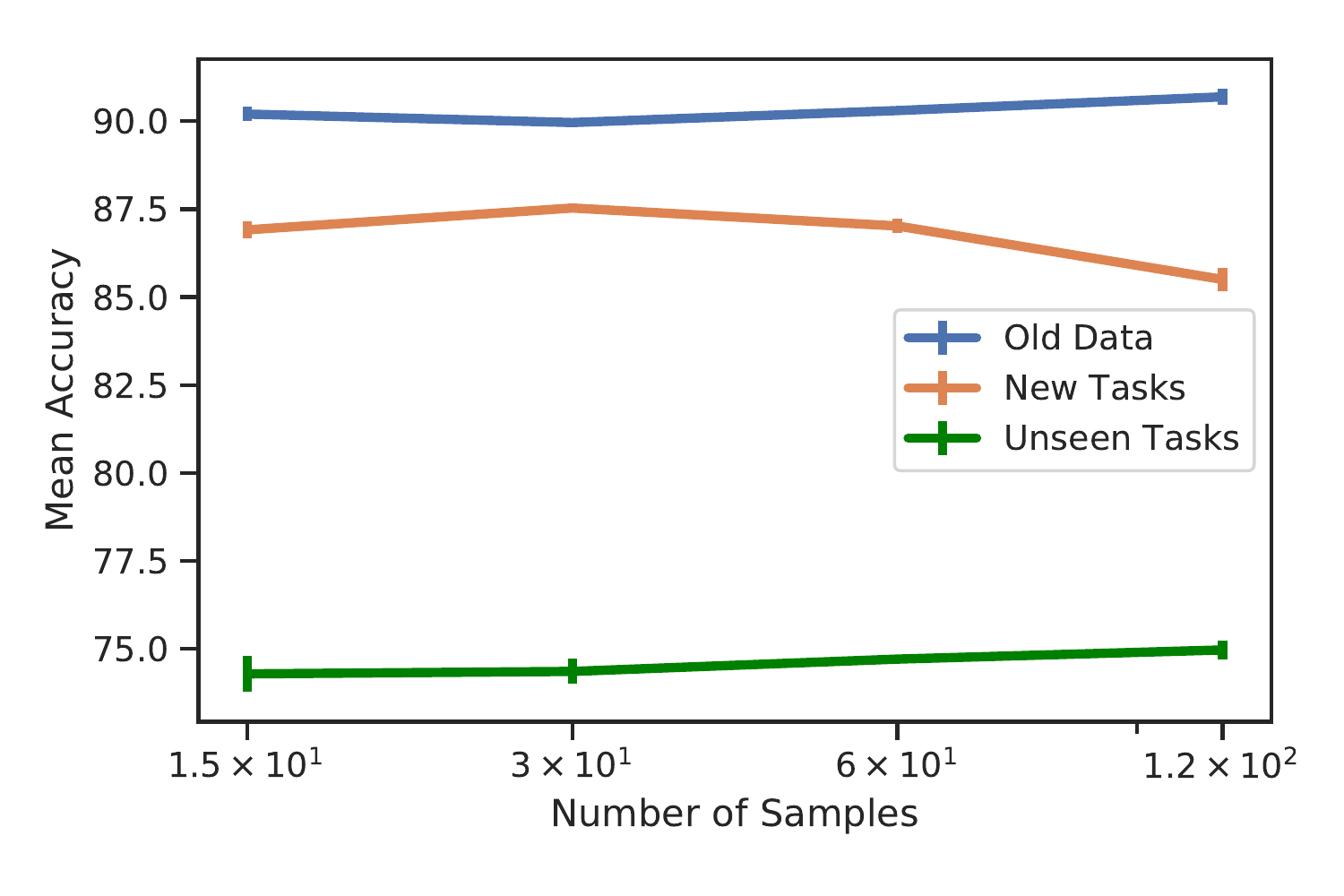}
\caption{Performance of the EIML model with different number of exemplars. The figure shows the mean accuracy averaged over 2000 episodes. The errorbars indicate the $95\%$ confidence interval.}
\label{Fig:sample_percentage}
\end{figure}

While using exemplars in EIML to retain information about the old tasks for meta-training our model incrementally, we kept the number of exemplars fixed to $15$. We investigate the effect of storing more exemplars per class to see if it helps  improve performance on old, new or unseen tasks. We investigated this in a \shot{5}{5} setup for MiniImageNet using PN with number of examples $\in$ $\{15, 30, 60, 120\}$. The outcome of this experiment is shown in Fig. \ref{Fig:sample_percentage}. This shows that increasing the number of exemplars per class does not yield much positive advantage on any of the task segments \ie old, new or unseen.

\section{Additional experimental results on \di}
In Tables \ref{Tab:mim51}, \ref{Tab:mim51++}, \ref{Tab:mim552}, \ref{Tab:mim512}, \ref{Tab:mim552++} \& \ref{Tab:mim512++} we report some additional experimental results on the \di dataset which explain the effects of domain gap better while using the baseline methods and our proposed algorithm. In our original paper, in Table 4 and Table 5, we showed the result of applying Incremental Meta-Learning (IML) on DomainImageNet for \shot{5}{5} setup using PN and ECM where the model is first trained using natural object classes and then incrementally trained using man-made object classes. Here we provide results for the same scenario in a \shot{1}{5} setup and also the results for PN and ECM in both \shot{1}{5} and \shot{1}{5} setup by reversing the domains \ie the model is first trained using tasks consisting of classes from man-made images and then incrementally trained using natural images.

\begin{table*}[t!]
\caption{Results of \shot{1}{5} classification accuracy of our method on different sets of \di using PN as the meta-learning algorithm. Here old tasks are from natural images and incremental tasks are from man-made images.}
\centering{
\small
\resizebox{0.85\textwidth}{!}{
\begin{tabular}{|l|ll|ll|l|}
\hline
                      \multicolumn{1}{|C{\colwidthA}|}{Model} & \multicolumn{1}{C{\colwidthA}}{Old classes from old domain (32)} & \multicolumn{1}{C{\colwidthA}|}{New classes from new domain (32)} & \multicolumn{1}{C{\colwidthA}}{Unseen classes from old domain (20)} & \multicolumn{1}{C{\colwidthA}|}{Unseen classes from new domain (20)} & \multicolumn{1}{C{\colwidthA}|}{Unseen classes from both domains (40)} \\
                       \hline

\rowcolor{orange!20} NU   & $56.32 \pm 0.50$    & $32.88 \pm 0.34$     & $37.90 \pm 0.37$     & $33.91 \pm 0.33$     & $39.56 \pm 0.40$     \\ \hline
FT     & $37.12 \pm 0.36$    & $\mathbf{61.34 \pm 0.56}$     & $32.81 \pm 0.32$     & $\mathbf{46.61 \pm 0.43}$     & $42.91 \pm 0.42$     \\
DFA     & $37.20 \pm 0.37$    & $60.23 \pm 0.55$     & $32.32 \pm 0.32$     & $46.12 \pm 0.44$     & $42.20 \pm 0.41$     \\
IDA   & $\mathbf{42.57 \pm 0.41}$    & $59.31 \pm 0.55$     & $\mathbf{35.01 \pm 0.35}$     & $45.55 \pm 0.43$     & $\mathbf{43.91 \pm 0.42}$     \\

\hline
\rowcolor{green!20} PAR & $62.23 \pm 0.50$    & $67.25 \pm 0.53$     & $40.06 \pm 0.39$     & $50.78 \pm 0.45$     & $54.57 \pm 0.48$          \\ \hline
\end{tabular}}
\label{Tab:mim51}
}
\end{table*}

\renewcommand{\colwidthA}{1.8cm}
\begin{table*}[t!]
\caption{Results of \shot{1}{5} classification accuracy of our method on different sets of \di using ECM as the meta-learning algorithm. Here old tasks are from natural images and incremental tasks are from man-made images.}
\centering{
\small
\resizebox{0.85\textwidth}{!}{
\begin{tabular}{|l|ll|ll|l|}
\hline
                      \multicolumn{1}{|C{\colwidthA}|}{Model} & \multicolumn{1}{C{\colwidthA}}{Old classes from old domain (32)} & \multicolumn{1}{C{\colwidthA}|}{New classes from new domain (32)} & \multicolumn{1}{C{\colwidthA}}{Unseen classes from old domain (20)} & \multicolumn{1}{C{\colwidthA}|}{Unseen classes from new domain (20)} & \multicolumn{1}{C{\colwidthA}|}{Unseen classes from both domains (40)} \\
                       \hline
\rowcolor{orange!20} NU  & $72.58 \pm 0.40$    & $39.68 \pm 0.37$     & $45.33 \pm 0.39$     & $40.43 \pm 0.34$     & $46.72 \pm 0.39$     \\ \hline
FT     & $48.54 \pm 0.40$    & $\mathbf{76.46 \pm 0.43}$     & $39.37 \pm 0.35$     & $\mathbf{54.93 \pm 0.41}$     & $51.64 \pm 0.43$     \\
DFA     & $49.83 \pm 0.40$    & $74.63 \pm 0.43$     & $40.36 \pm 0.36$     & $54.80 \pm 0.41$     & $52.32 \pm 0.43$     \\
IDA    & $\mathbf{69.82 \pm 0.41}$    & $62.56 \pm 0.44$     & $\mathbf{46.22 \pm 0.39}$     & $49.83 \pm 0.40$     & $\mathbf{53.03 \pm 0.41}$     \\ \hline

\rowcolor{green!20} PAR & $70.10 \pm 0.42$    & $73.70 \pm 0.44$     & $46.81 \pm 0.40$     & $55.19 \pm 0.41$     & $57.97 \pm 0.42$          \\ \hline
\end{tabular}}

\label{Tab:mim51++}
}
\end{table*}

\renewcommand{\colwidthA}{1.8cm}
\begin{table*}[t!]
\caption{Results of \shot{5}{5} classification accuracy of our method on different sets of \di  using PN as the meta-learning algorithm. Here old tasks are from man-made images and incremental tasks are from natural images.}
\centering{
\small
\resizebox{0.85\textwidth}{!}{
\begin{tabular}{|l|ll|ll|l|}
\hline
                      \multicolumn{1}{|C{\colwidthA}|}{Model} & \multicolumn{1}{C{\colwidthA}}{Old classes from old domain (32)} & \multicolumn{1}{C{\colwidthA}|}{New classes from new domain (32)} & \multicolumn{1}{C{\colwidthA}}{Unseen classes from old domain (20)} & \multicolumn{1}{C{\colwidthA}|}{Unseen classes from new domain (20)} & \multicolumn{1}{C{\colwidthA}|}{Unseen classes from both domains (40)} \\
                       \hline
\rowcolor{orange!20} NU   & $91.75 \pm 0.18$    & $50.67 \pm 0.37$     & $70.94 \pm 0.35$     & $45.04 \pm 0.34$     & $62.57 \pm 0.44$     \\ \hline
FT     & $71.41 \pm 0.31$    & $\mathbf{84.40 \pm 0.30}$     & $57.11 \pm 0.32$     & $59.61 \pm 0.43$     & $63.40 \pm 0.38$     \\
DFA     & $73.51 \pm 0.31$    & $83.31 \pm 0.31$     & $58.87 \pm 0.32$     & $59.64 \pm 0.42$     & $64.36 \pm 0.37$     \\
IDA   & $\mathbf{88.39 \pm 0.23}$    & $81.24 \pm 0.34$     & $\mathbf{70.83 \pm 0.34}$     & $\mathbf{60.82 \pm 0.39}$     & $\mathbf{71.13 \pm 0.38}$     \\

\rowcolor{green!20} PAR & $90.35 \pm 0.21$    & $87.11 \pm 0.27$     & $73.03 \pm 0.34$     & $62.44 \pm 0.42$     & $75.27 \pm 0.38$          \\ \hline
\end{tabular}}
\label{Tab:mim552}
}
\end{table*}

\renewcommand{\colwidthA}{1.8cm}
\begin{table*}[t!]
\caption{Results of \shot{1}{5} classification accuracy of our method on different sets of \di using PN as the meta-learning algorithm. Here old tasks are from man-made images and incremental tasks are from natural images.}
\centering{
\small
\resizebox{0.85\textwidth}{!}{
\begin{tabular}{|l|ll|ll|l|}
\hline
                      \multicolumn{1}{|C{\colwidthA}|}{Model} & \multicolumn{1}{C{\colwidthA}}{Old classes from old domain (32)} & \multicolumn{1}{C{\colwidthA}|}{New classes from new domain (32)} & \multicolumn{1}{C{\colwidthA}}{Unseen classes from old domain (20)} & \multicolumn{1}{C{\colwidthA}|}{Unseen classes from new domain (20)} & \multicolumn{1}{C{\colwidthA}|}{Unseen classes from both domains (40)} \\
                       \hline
\rowcolor{orange!20} NU   & $72.45 \pm 0.51$    & $33.41 \pm 0.34$     & $51.48 \pm 0.48$     & $32.32 \pm 0.31$     & $43.88 \pm 0.45$     \\ \hline
FT     & $46.08 \pm 0.42$    & $\mathbf{61.72 \pm 0.56}$     & $39.90 \pm 0.39$     & $42.38 \pm 0.43$     & $44.85 \pm 0.43$     \\
DFA     & $48.35 \pm 0.44$    & $58.13 \pm 0.56$     & $40.66 \pm 0.40$     & $41.30 \pm 0.41$     & $44.04 \pm 0.42$     \\
IDA & $\mathbf{58.98 \pm 0.47}$    & $58.90 \pm 0.56$     & $\mathbf{46.33 \pm 0.44}$     & $\mathbf{43.14 \pm 0.42}$     & $\mathbf{48.33 \pm 0.44}$     \\

\hline
\rowcolor{green!20} PAR & $73.31 \pm 0.50$    & $68.54 \pm 0.55$     & $52.29 \pm 0.45$     & $44.28 \pm 0.45$     & $58.11 \pm 0.50$          \\ \hline
\end{tabular}}
\label{Tab:mim512}
}
\end{table*}

\renewcommand{\colwidthA}{1.8cm}
\begin{table*}[t!]
\caption{Results of \shot{5}{5} classification accuracy of our method on different sets of \di using ECM as the meta-learning algorithm. Here old tasks are from man-made images and incremental tasks are from natural images.}
\centering{
\small
\resizebox{0.85\textwidth}{!}{
\begin{tabular}{|l|ll|ll|l|}
\hline
                      \multicolumn{1}{|C{\colwidthA}|}{Model} & \multicolumn{1}{C{\colwidthA}}{Old classes from old domain (32)} & \multicolumn{1}{C{\colwidthA}|}{New classes from new domain (32)} & \multicolumn{1}{C{\colwidthA}}{Unseen classes from old domain (20)} & \multicolumn{1}{C{\colwidthA}|}{Unseen classes from new domain (20)} & \multicolumn{1}{C{\colwidthA}|}{Unseen classes from both domains (40)} \\
                       \hline
\rowcolor{orange!20} NU   & $59.58 \pm 0.40$    & $91.91 \pm 0.16$     & $73.04 \pm 0.32$     & $53.20 \pm 0.37$     & $68.12 \pm 0.40$     \\ \hline
FT     & $\mathbf{87.66 \pm 0.24}$    & $72.11 \pm 0.31$     & $61.39 \pm 0.31$     & $\mathbf{63.87 \pm 0.40}$     & $67.91 \pm 0.35$     \\
DFA     & $82.45 \pm 0.31$    & $77.01 \pm 0.29$     & $62.91 \pm 0.32$     & $62.31 \pm 0.42$     & $67.79 \pm 0.36$     \\
IDA    & $79.80 \pm 0.33$    & $\mathbf{90.43 \pm 0.19}$     & $\mathbf{74.04 \pm 0.31}$     & $61.31 \pm 0.39$     & $\mathbf{72.60 \pm 0.37}$     \\ \hline
\rowcolor{green!20} PAR & $85.66 \pm 0.28$    & $89.44 \pm 0.20$     & $74.60 \pm 0.32$     & $66.12 \pm 0.39$     & $76.92 \pm 0.35$          \\ \hline
\end{tabular}}

\label{Tab:mim552++}
}
\end{table*}

\renewcommand{\colwidthA}{1.8cm}
\begin{table*}[t!]
\caption{Results of \shot{1}{5} classification accuracy of our method on different sets of \di using ECM as the meta-learning algorithm. Here old tasks are from man-made images and incremental tasks are from natural images.}
\centering{
\small
\resizebox{0.85\textwidth}{!}{
\begin{tabular}{|l|ll|ll|l|}
\hline
                      \multicolumn{1}{|C{\colwidthA}|}{Model} & \multicolumn{1}{C{\colwidthA}}{Old classes from old domain (32)} & \multicolumn{1}{C{\colwidthA}|}{New classes from new domain (32)} & \multicolumn{1}{C{\colwidthA}}{Unseen classes from old domain (20)} & \multicolumn{1}{C{\colwidthA}|}{Unseen classes from new domain (20)} & \multicolumn{1}{C{\colwidthA}|}{Unseen classes from both domains (40)} \\
                       \hline
\rowcolor{orange!20} NU   & $42.27 \pm 0.38$    & $79.97 \pm 0.40$     & $53.49 \pm 0.42$     & $37.67 \pm 0.34$     & $49.74 \pm 0.42$ \\ \hline
FT     & $\mathbf{73.37 \pm 0.47}$    & $51.59 \pm 0.40$     & $42.22 \pm 0.35$     & $\mathbf{47.09 \pm 0.42}$     & $48.64 \pm 0.41$     \\
DFA     & $66.49 \pm 0.50$    & $56.71 \pm 0.42$     & $43.68 \pm 0.36$     & $46.29 \pm 0.43$     & $49.12 \pm 0.41$     \\
IDA    & $60.64 \pm 0.46$    & $\mathbf{76.91 \pm 0.41}$     & $\mathbf{54.02 \pm 0.42}$     & $44.01 \pm 0.39$     & $\mathbf{53.16 \pm 0.42}$     \\ \hline
\rowcolor{green!20} PAR & $68.86 \pm 0.47$    & $73.87 \pm 0.43$     & $53.91 \pm 0.42$     & $48.00 \pm 0.42$     & $58.00 \pm 0.43$          \\ \hline
\end{tabular}}
\label{Tab:mim512++}
}
\end{table*}

\section{Meta-Learning algorithms and their dependency on the number of shots and ways} \label{sec:lifting}

One of the main limitations of the simplest implementation of our method, which is based on PN, is the fact that old and new tasks must have the same number of classes. This is not a limitation of our method, but of the use of PN as a base learner. Our second implementation, which uses ECM instead of PN, is not subject to this restriction. Therefore, in  \tabref{tab:PNsweep} and \tabref{tab:ECMsweep} we show the dependency of our method on the number of shots and ways as they vary across old and new datasets. We first see that PN is highly dependent on the number of shots and ways it is being trained on. We see a drop in performance when the shots are changed between meta-training and few-shot testing. We also notice for PN, training using a larger number of ways helps to improve performance. However, we notice that for ECM, the performance remains the same irrespective of what configuration the \ml model was trained with.  These results use the feature extractor proposed in the original PN \cite{snell2017prototypical} paper instead of ResNet-12 for simplicity and ease of reproducibility. All models were trained using identical hyperparameter settings (batch size, learning rate, optimization scheme). These results collectively show that our method is flexible and allows meta-training with an arbitrary number of ways and shots. The results also show that ECM is a more practical choice of base learner, despite the increased training cost, since it provides flexibility to be be used with any number of shots and ways, without having to train separate models. 

\renewcommand{\colwidthA}{1.8cm}
\begin{table*}[t!]
\centering
\caption{Performance of different models trained using PN \cite{snell2017prototypical} on different few-shot sets with varying ``ways'' and ``shots''. Mean accuracy averaged over 2000 episodes is shown here.}
\begin{tabular}{|l|lll|lll|}
\hline
\multirow{2}{*}{Model} & \multicolumn{3}{c|}{1-shot}                                                & \multicolumn{3}{c|}{5-shot}                                                \\
\cline{2-7}
                       & \multicolumn{1}{c}{5-way} & \multicolumn{1}{c}{10-way} & \multicolumn{1}{c|}{20-way} & \multicolumn{1}{c}{5-way} & \multicolumn{1}{c}{10-way} & \multicolumn{1}{c|}{20-way}                     \\
                       
                       \hline
 \shot{1}{5} & 49.35   &     33.78  &   21.99 &    65.63    &     49.60 & 35.76         \\
 \shot{1}{10} & 51.47   &     35.65  &    23.54 &    68.45   &     52.80 & 38.93        \\
  \shot{5}{5} & 47.09  &     31.94  &    20.68 &    69.09    &     53.52 & 39.78       \\    
  \shot{5}{10} & 45.83  &     30.89  &    19.90 &    69.93    &     54.77 & 41.05        \\    
 \hline
 Range & 5.63 &	4.76 &	3.64 & 4.30 &	5.18	&5.29\\
 \hline
\end{tabular}
\label{tab:PNsweep}
\end{table*}

\renewcommand{\colwidthA}{1.8cm}
\begin{table*}[t!]
\centering
\caption{Performance of different models trained using ECM \cite{ravichandran2019few} on different few-shot sets with varying ``ways'' and ``shots''. Mean accuracy averaged over 2000 episodes is shown here.}
\begin{tabular}{|l|lll|lll|}
 \hline
\multirow{2}{*}{Model} & \multicolumn{3}{c|}{1-shot}                                                & \multicolumn{3}{c|}{5-shot}                                                \\
\cline{2-7}
                       & \multicolumn{1}{c}{5-way} & \multicolumn{1}{c}{10-way} & \multicolumn{1}{c|}{20-way} & \multicolumn{1}{c}{5-way} & \multicolumn{1}{c}{10-way} & \multicolumn{1}{c|}{20-way}                     \\
                       \hline
 \shot{1}{5} & 50.13   &     35.15  &   23.48 &    68.40    &     53.30 &39.68         \\
 \shot{1}{10} & 50.91   &     35.69  &    23.84 &    69.62    &     54.34 & 40.53        \\
  \shot{5}{5} & 51.15  &     35.95  &    24.08 &    69.50   &    54.40 & 40.96        \\    
  \shot{5}{10} & 50.78  &    35.44  &    23.73 &    69.64   &     54.46 & 40.72       \\    
 \hline 
 Range & 1.0224 & 0.7921 & 0.6009 & 1.2455 & 1.1688 & 1.2748 \\
 \hline
\end{tabular}
\label{tab:ECMsweep}
\end{table*}

\end{document}


\newcommand{\wraptxt}[1]{#1\xspace}
\newcommand{\bftab}{\fontseries{b}\selectfont}
\newcommand{\figref}[1]{\wraptxt{Figure \ref{#1}}}
\newcommand{\secref}[1]{\wraptxt{Section \ref{#1}}}
\newcommand{\tabref}[1]{\wraptxt{Table \ref{#1}}}
\newcommand{\ds}{\wraptxt{dataset}}
\newcommand{\ml}{\wraptxt{meta-learning}}
\newcommand{\mt}{\wraptxt{meta-training}}
\newcommand{\mml}{\wraptxt{metric-based meta-learning}}
\newcommand{\iml}{\wraptxt{incremental meta-learning}}
\newcommand{\fs}{\wraptxt{few-shot}}
\newcommand{\ct}{\wraptxt{catastrophic forgetting}}
\newcommand\numberthis{\addtocounter{equation}{1}\tag{\theequation}}
\newcolumntype{C}[1]{>{\centering\let\newline\arraybackslash\hspace{0pt}}m{#1}}
\newcommand{\setB}{\wraptxt{\mathcal{B}}}
\newcommand{\setC}{\wraptxt{\mathcal{C}}}
\newcommand{\setT}{\wraptxt{\mathcal{T}}}

\newcommand{\mi}{\wraptxt{miniImageNet}}
\newcommand{\ti}{\wraptxt{tieredImageNet}}
\newcommand{\di}{\wraptxt{domainImageNet}}

\newcommand{\ie}{\wraptxt{i.e.,}}

\newcommand{\shot}[2]{#1-shot #2-way}

\renewcommand{\note}[1]{{\color{red}#1}}
\newcommand{\notea}[1]{{\color{blue}#1}}

\pagestyle{headings}
\mainmatter
\def\ECCVSubNumber{100}  

\title{Author Guidelines for ECCV Submission} 

\titlerunning{ECCV-20 submission ID \ECCVSubNumber} 
\authorrunning{ECCV-20 submission ID \ECCVSubNumber} 
\author{Anonymous ECCV submission}
\institute{Paper ID \ECCVSubNumber}

\title{Incremental Meta-Learning} 

\maketitle

\section{Experiments}
\subsection{\mi Experiments}
\newcommand{\colwidthA}{1.8cm}
\begin{table*}[t!]
\centering{
\small
\resizebox{0.95\textwidth}{!}{
\begin{tabular}{l|lll|lll|l}
\hline
\multirow{2}{*}{Model} & \multicolumn{3}{c|}{\shot{1}{5}} & \multicolumn{3}{c|}{\shot{1}{5}} & \multirow{2}{*}{Base data} \\
\cline{2-7}
                       & \multicolumn{1}{C{\colwidthA}}{Base classes (32)} & \multicolumn{1}{C{\colwidthA}}{Incremental classes (32) } & \multicolumn{1}{C{\colwidthA}|}{Meta-test (20)} & \multicolumn{1}{C{\colwidthA}}{Base classes} & \multicolumn{1}{C{\colwidthA}}{Incremental classes } & \multicolumn{1}{c|}{Meta-test} &                \\ 
                       \hline
Base   & $73.84 \pm 0.50$    & $49.05 \pm 0.48$    & $50.55 \pm 0.42$ & $91.17 \pm 0.18$  & $68.35 \pm 0.39$ & $68.60 \pm 0.33$ & $-$              \\ 
\hline
Continued    & $60.64 \pm 0.49$    & $72.61 \pm 0.51$     & $53.60 \pm 0.42$ & $82.25 \pm 0.26$  & $89.63 \pm 0.22$ & $72.13 \pm 0.33$ & $0$                \\
Distillation      & $60.77 \pm 0.49$    & $72.23 \pm 0.51$     & $53.81 \pm 0.42$ & $82.53 \pm 0.26$  & $89.32 \pm 0.22$ & $72.07 \pm 0.33$ & $0$                \\
Identity    & $66.54 \pm 0.49$    & $71.92 \pm 0.51$     & $54.86 \pm 0.42$ & $89.14 \pm 0.21$  & $87.32 \pm 0.25$ & $75.11 \pm 0.31$ &  $1/600$            \\
Exemplar & $68.95 \pm 0.50$    & $71.43 \pm 0.52$     & $55.52 \pm 0.43$ & $90.20 \pm 0.20$  & $86.91 \pm 0.25$ & $74.39 \pm 0.50$ &  $15/600$           \\ \hline
Full         & $74.65 \pm 0.49$    & $75.85 \pm 0.50$     & $56.88 \pm 0.43$ & $91.77 \pm 0.17$  & $92.49 \pm 0.17$ & $74.80 \pm 0.31$ &  $600/600$          \\ \hline
\end{tabular}}
\caption{Results of classification accuracy on different sets of \mi using our distillation method and meta-training using Prototypical networks \cite{snell2017prototypical}. The evaluation is performed on 2000 episodes and we report the mean and the $95\%$ confidence interval. We report the performance of  models trained on the base classes, the model we obtain by continued training, the model obtained by feature distillation, model using class identities and that using exemplar data. We show this performance on three different sets: tasks sampled from the base classes, from  incremental classes and tasks sampled from the meta-test sets. }
\label{Tab:miproto}
}
\end{table*}

\begin{table*}[t!]
\centering{
\small
\resizebox{0.95\textwidth}{!}{
\begin{tabular}{l|lll|lll|l}
\hline
\multirow{2}{*}{Model} & \multicolumn{3}{c|}{\shot{1}{5}} & \multicolumn{3}{c|}{\shot{1}{5}} & \multirow{2}{*}{Base data} \\
\cline{2-7}
                       & \multicolumn{1}{C{\colwidthA}}{Base classes (176)} & \multicolumn{1}{C{\colwidthA}}{Incremental classes (175) } & \multicolumn{1}{C{\colwidthA}|}{Meta-test (160)} & \multicolumn{1}{C{\colwidthA}}{Base classes} & \multicolumn{1}{C{\colwidthA}}{Incremental classes } & \multicolumn{1}{c|}{Meta-test} &                \\
                       \hline
Base   & $73.82 \pm 0.43$    & $53.00 \pm 0.43$     & $52.77 \pm 0.37$ & $89.38 \pm 0.38$  & $71.90 \pm 0.36$ & $71.57 \pm 0.36$ & $-$                 \\ \hline
Continued    & $63.71 \pm 0.43$    & $75.05 \pm 0.43$     & $56.00 \pm 0.38$ & $82.90 \pm 0.21$  & $89.37 \pm 0.21$ & $74.29 \pm 0.32$ & $0$                 \\
Distillation      & $64.66 \pm 0.42$    & $75.71 \pm 0.43$     & $56.68 \pm 0.39$ & $83.37 \pm 0.21$  & $89.70 \pm 0.21$ & $74.69 \pm 0.31$ & $0$          \\
Identity    & $72.52 \pm 0.42$    & $68.43 \pm 0.44$     & $57.13 \pm 0.39$ & $88.46 \pm 0.27$  & $85.45 \pm 0.27$ & $75.55 \pm 0.30$ &  $1/600$             \\\hline
Full        & $74.40 \pm 0.40$    & $75.74 \pm 0.42$     & $59.02 \pm 0.39$ & $89.68 \pm 0.21$  & $89.93 \pm 0.21$ & $77.60 \pm 0.30$ &  $600/600$          \\ \hline
\end{tabular}}
\caption{Results of classification accuracy on different sets of  \mi using our method and meta-training using Shot Invariant \cite{ravichandran2019few}.  The evaluation is performed on 2000 episodes and we report the mean and the $95\%$ confidence interval. We report the performance of  models trained on the base classes, the model we obtain by continued training, the model obtained by feature distillation, model using class identities and that using exemplar data. We show this performance on three different sets: tasks sampled from the base classes, from  incremental classes and tasks sampled from the meta-test sets.}
\label{Tab:miprotopp}
}
\end{table*}

\pagebreak

\subsection{\ti Experiments}

\begin{table*}[t!]
\centering{
\small
\resizebox{0.95\textwidth}{!}{
\begin{tabular}{l|lll|lll|l}
\hline
\multirow{2}{*}{Model} & \multicolumn{3}{c|}{\shot{1}{5}} & \multicolumn{3}{c|}{\shot{1}{5}} & \multirow{2}{*}{Base data} \\
\cline{2-7}
                       & \multicolumn{1}{C{\colwidthA}}{Base classes} & \multicolumn{1}{C{\colwidthA}}{Incremental classes } & \multicolumn{1}{C{\colwidthA}|}{Meta-test} & \multicolumn{1}{C{\colwidthA}}{Base classes} & \multicolumn{1}{C{\colwidthA}}{Incremental classes } & \multicolumn{1}{c|}{Meta-test} &                \\
                       \hline
Base   & $73.10 \pm 0.52$    & $66.18 \pm 0.43$     & $56.82 \pm 0.50$ & $89.03 \pm 0.27$  & $81.97 \pm 0.37$ & $75.78 \pm 0.43$ & $-$                \\ \hline
Continued    & $71.87 \pm 0.52$    & $71.03 \pm 0.52$     & $58.63 \pm 0.50$ & $87.77 \pm 0.29$  & $87.60 \pm 0.30$ & $78.20 \pm 0.42$ & $0$                \\
Distillation      & $72.03 \pm 0.51$    & $70.83 \pm 0.53$     & $58.81 \pm 0.50$ & $87.82 \pm 0.29$  & $87.38 \pm 0.30$ & $78.11 \pm 0.42$ & $0$                \\
Identity    & $72.65 \pm 0.51$    & $70.17 \pm 0.53$     & $58.71 \pm 0.50$ & $89.03 \pm 0.27$  & $86.91 \pm 0.31$ & $78.40 \pm 0.42$ &  $1/600$            \\ \hline
Full         & $78.57 \pm 0.51$    & $77.43 \pm 0.50$     & $61.87 \pm 0.51$ & $91.05 \pm 0.24$  & $90.44 \pm 0.26$ & $80.58 \pm 0.40$ &  $600/600$          \\ \hline
\end{tabular}}
\caption{Results of classification accuracy on different sets of \ti using our distillation method and meta-training using Prototypical networks \cite{snell2017prototypical}. The evaluation is performed on 2000 episodes and we report the mean and the $95\%$ confidence interval. We report the performance of  models trained on the base classes, the model we obtain by continued training, the model obtained by feature distillation, model using class identities and that using exemplar data. We show this performance on three different sets: tasks sampled from the base classes, from  incremental classes and tasks sampled from the meta-test sets. }
\label{Tab:tiproto}
}
\end{table*}

\section{Experiments}

\subsection{Quantitative Results}
We evaluate our methods on the standard \shot{1}{5} and \shot{5}{5} setup by sampling tasks consisting of classes from the held-out test set, base set and the incremental set. As opposed to the standard practice found in few-shot learning literature, we are not only interested in performance on the held-out test set but also on the classes from the base train set as well as the incremental train set to measure \ct. Therefore, we report numbers on the test set, base set and the incremental set to demonstrate performance of the model on novel classes as well as previously seen classes. The result on ``Base classes'' (randomly sampled episodes from classes present in base set) reflects how much information the model retains from its base training, while the result on ``Incremental classes'' (randomly sampled episodes from classes present in incremental set) illustrates how much new knowledge is obtained during incremental training. The result on ``Meta test'' shows how the model generalizes to unseen classes from the test set and how much its meta-learning ability gets enhanced when compared to meta-train with only ``Base classes''.
We compare the two variants of our incremental meta-training methods with four baselines namely: base model, continued training, feature distillation and full training.

\paragraph{\mi:} We show the results of using incremental meta-learning with Prototypical Networks as the meta-learning algorithm in \tabref{Tab:miproto} and using \cite{ravichandran2019few} in \tabref{Tab:miprotopp}. From this table we notice that continued training and feature distillation diminish the performance on base classes as compared to the base model for both \shot{1}{5} and \shot{5}{5}. While we expect this for continued training due to \ct, we notice a similar effect for feature distillation too. Additionally, this is true for both the meta-learning algorithms.
We notice that our proposed methods diminish the performance on incremental classes as compared to continued training and feature distillation and this is primarily because continued training and feature distillation methods majorly focus on the incremental classes and this gain comes at the cost of performance on the base classes. This is consistent with both the meta-learning algorithms and the different shot and way setting we have used.
Incremental training using any method improves the performance on both incremental classes as well as test classes when compared to the base model.
Overall, we empirically observed that incremental training by retaining class identifies performs better than incremental training using exemplars. Hence for the rest of the experiments, we focused on this method.


\paragraph{\ti:} In \tabref{Tab:tiproto}, we show the results of our method for meta-learning using \cite{snell2017prototypical} on \ti. From this table, we notice that the improvement by using more classes (base vs full) is relatively smaller compared to \mi. When  the number of classes used for base training is high, the generalization ability of the network is better. Consequently, we see very little \ct and increase in meta-learning performance. We also see the that our proposed method and the baselines are very similar. This further strengthens our argument for the lack of \ct in this \ds. This raises the question if incremental learning is always effective during meta-training and what type of new classes could potentially help in improving the performance of incremental training. Our experiments on \di attempt to answer this question.

\paragraph{\di:} The result of meta-training and testing using a \shot{5}{5} on \di is shown in Table \ref{Tab:mim55} and also in Table \ref{Tab:mim55++}, \ref{Tab:mim51}, \ref{Tab:mim51++}, \ref{Tab:mim552}, \ref{Tab:mim512}, \ref{Tab:mim552++}, \ref{Tab:mim512++} in the supplementary material. Here, we base train using classes that belong natural objects and incrementally train using classes that belong to man-made objects and vice-versa. Hence, there is significant domain shift between base training and incremental training. We test on three different sets: unseen classes from natural objects, unseen classes from man-made objects and unseen classes from a mix of both. We show the results when we use the natural images as a base domain and incremental domain as man-made classes in \tabref{Tab:mim55}, \tabref{Tab:mim55++}, \tabref{Tab:mim51} and \tabref{Tab:mim51++}. Furthermore, we show results when we use man-made classes as base domain and natural classes as incremental domain in \tabref{Tab:mim552}, \tabref{Tab:mim512}, \tabref{Tab:mim552++} and \tabref{Tab:mim512++}.

These tables show that the accuracy on the joint test set improves significantly when compared to the base model. But if we analyze the performance for seen and unseen domains, we see that most of the gain is for the unseen domain \ie man-made objects. Also, \ct is quite large here since there is a significant domain shift from between the base and incremental classes for the baseline methods. This effect is also seen on testing with unseen classes on the same domain. We see that our proposed method is able to show improvement across the board when compared to the base model. The numbers for \shot{1}{5} way and the results using the reverse domain training \ie base train using man-made objects and incremental test using natural objects on the 3 sets for the incremental meta-learning algorithms show similar trends. This suggests what is used for incremental training matters, specifically adding classes with sufficiently different image level statistics is the best way to obtain improvements. Adding unseen classes when they belong to the same domain marginally improves the performance. This might seem intuitive and our experimental results reaffirm this. While this trend is expected for samples belonging to the same classes, we surprisingly find it to be same for samples belonging to unseen classes from the same domain. Our method is successfully able to mitigate \ct to a large extent and perform well across different domains. 

\begin{table*}[t!]
\centering{
\small
\resizebox{0.95\textwidth}{!}{
\begin{tabular}{l|lllll}
\hline
\cline{2-6}
                      \multicolumn{1}{C{\colwidthA}}{Model} & \multicolumn{1}{C{\colwidthA}}{Seen classes from base domain} & \multicolumn{1}{C{\colwidthA}}{Seen classes from incremental domain} & \multicolumn{1}{C{\colwidthA}|}{Unseen classes from base domain} & \multicolumn{1}{C{\colwidthA}}{Unseen classes from incremental domain} & \multicolumn{1}{C{\colwidthA}}{Unseen classes from both domains} &                \\
                       \hline
Base   & $86.94 \pm 0.22$    & $49.14 \pm 0.36$     & $57.66 \pm 0.38$     & $51.72 \pm 0.32$     & $59.59 \pm 0.35$ \\ \hline
Continued    & $64.42 \pm 0.35$    & $84.80 \pm 0.28$     & $50.72 \pm 0.38$     & $71.16 \pm 0.32$     & $65.44 \pm 0.40$ \\
Distillation      & $65.12 \pm 0.35$    & $83.95 \pm 0.29$     & $51.33 \pm 0.38$     & $70.46 \pm 0.33$     & $65.52 \pm 0.40$ \\
Identity    & $81.26 \pm 0.27$    & $82.06 \pm 0.30$     & $59.32 \pm 0.39$     & $70.61 \pm 0.32$     & $70.36 \pm 0.36$ \\ \hline
Full & $87.44 \pm 0.22$    & $88.77 \pm 0.25$     & $58.59 \pm 0.37$     & $74.46 \pm 0.32$     & $74.02 \pm 0.37$ \\ \hline
\end{tabular}}
\caption{Results of 5-shot 5-way classification accuracy of our method on different sets of \di  using \cite{snell2017prototypical} as meta-learning algorithm. The evaluation is performed on  2000 episodes and we report the mean and the $95\%$ confidence interval.}
\label{Tab:mim55}
}
\end{table*}

\subsection{Additional Experiments}
\begin{figure}[h]
\centering
\includegraphics[width=8cm]{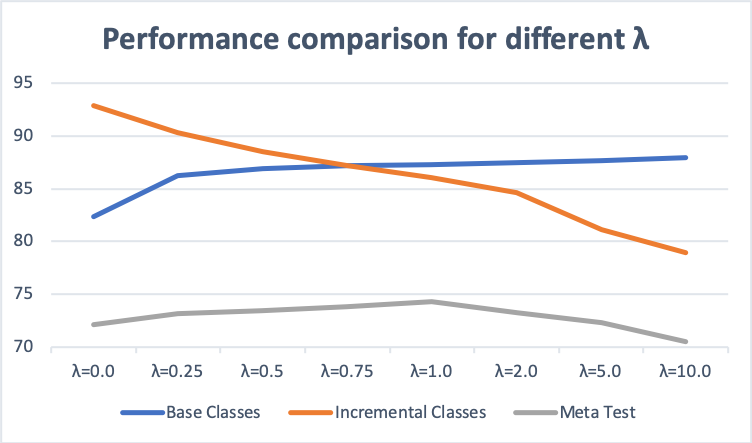}
\caption{Performance of the model's meta-learning ability on base, incremental and test set with $\lambda$ varying from $0.0$ to $10.0$. Performance on base class improves when $\lambda$ increases while it worsens the performance on incremental set. Best performance is obtained roughly in the middle, around $1.0$. \note{it would be best to show errorbars across multiple runs}}
\vspace{-0.1cm} \label{Fig:lambda_percentage}
\end{figure}

\paragraph{Varying $\lambda$ while training with base class-identity functions:}
When we retain the class-identity function for each of the base classes, we can  maintain a trade-off between the knowledge distillation loss and the standard meta-learning loss. This is controlled by the $\lambda$ parameter as specified in eq. 7. In this experiment, we investigate how changing the value of $\lambda$ affects the model's meta-learning ability on base, incremental and the meta-test set. We investigated this in  the \shot{5}{5} setup for \mi using \cite{snell2017prototypical}. We chose $\lambda \in \{0.0, 0.25, 0.5, 0.75, 1.0, 2.0, 5.0, 10.0\}$. In our original experiments, we used $\lambda=1.0$. Outcome of this experiment can be visualized in \figref{Fig:lambda_percentage} where we can see that as $\lambda$ is increased, the model's performance on base set improves while its performance deteriorates on the incremental set. This is intuitive as \note{the (there are a few articles missing in this section)} knowledge distillation loss adds additional constraint for the incremental model to be similar to the model learnt on base set and increasing its contribution in the overall loss enhances model's ability to perform better on base set. However, for the same reason, it diminishes the impact of standard meta-learning loss on the overall loss and hence, performance on incremental set gets diminished. At $\lambda=0.0$, the model only trains with the standard meta-learning loss and its performance is identical to our baseline named ``Continued Training''. At $\lambda$ value of $10.0$, knowledge distillation loss is dominating and the performance is similar (not fully identical though) to the model trained only using base set. While measuring performance on the test set, we observed that the ideal value of $\lambda$ lies somewhat between the values of $\lambda$ that provides best performance on the base set and the values of $\lambda$ that work best for the incremental set. 

\begin{figure}[ht]
\centering
\includegraphics[width=8cm]{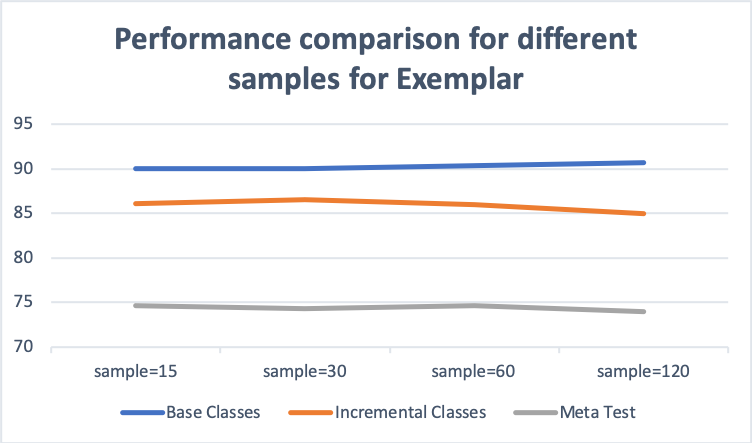}
\caption{Performance of the model on the number of exemplars kept while meta-training was done using base exemplars. Increasing number of exemplars to retain did not yield any positive advantange. \note{errorbars. Where are the horizontal marks? Why not show them?}}
\vspace{-0.1cm} \label{Fig:sample_percentage}
\end{figure}

\paragraph{Varying number of samples while training with base exemplars:}
While using base exemplars to retain information about the base classes for meta-training our model incrementally, we kept the number of exemplars fixed to $15$. We investigate the effect of storing more exemplars per class and see if it helps  to avoid \ct on the base set. We again investigated this in the  \shot{5}{5} setup for \mi using \cite{snell2017prototypical}. We chose the number of samples as $\{15, 30, 60, 120\}$. Outcome of this experiment is depicted in \figref{Fig:sample_percentage} which shows  that increasing the number of exemplars per class does not yield much positive advantage on either of the segments \ie base, incremental or the meta test-set. This is in accordance with our observations on others \ds as well that retaining more exemplars from base set does not help much and retaining only an identity representation per class can be sufficient while performing incremental meta-learning without incurring \ct to a large extent.

\section{Conclusion}
In this paper, we proposed an algorithm for incremental learning within a \ml setup. Incremental learning during meta-training provides a more realistic scenario for real-world few-shot applications where more and more rare classes are seen over time. To the best of our knowledge, we are the first to propose such an experimental setup using existing few-shot classification datasets. We show empirical results on standard benchmark datasets and show that our algorithm is very close to the performance obtained by using all the data at once. We also suggest additional baselines to test incremental learning for \mml which is robust to \ct. We would like to point out that our algorithm is quite generic and can work in conjunction with multiple existing \ml algorithms.

{\small
\bibliographystyle{ieee_fullname}
\balance
\bibliography{egbib}
}

\clearpage
\onecolumn
\section*{\hfil Supplementary Materials \hfil}
\maketitle
Here we report some additional experimental results on the \di dataset which explain the effects of domain gap better while using the baseline methods and our proposed algorithm.
\renewcommand{\colwidthA}{2.0cm}
\begin{table*}[h]
\centering{
\small
\begin{tabular}{lccccc}
\hline
             & Seen classes from base domain     & Seen classes from incremental domain     &  Unseen classes from  base domain       & Unseen classes from incremental Domain      & Unseen classes from both domains  \\ \hline
Base   & $87.86 \pm 0.20$    & $56.71 \pm 0.39$     & $63.30 \pm 0.38$     & $58.10 \pm 0.35$     & $66.09 \pm 0.35$  \\ \hline
Continued     & $67.35 \pm 0.34$    & $89.68 \pm 0.20$     & $55.37 \pm 0.38$     & $74.00 \pm 0.31$     & $69.98 \pm 0.39$  \\
Distillation      & $69.33 \pm 0.33$    & $88.72 \pm 0.22$     & $57.06 \pm 0.38$     & $73.97 \pm 0.31$     & $70.77 \pm 0.38$  \\
Identity    & $86.09 \pm 0.22$    & $81.82 \pm 0.28$     & $64.22 \pm 0.38$     & $69.92 \pm 0.33$     & $72.64 \pm 0.33$  \\ \hline
Full & $86.83 \pm 0.22$    & $88.84 \pm 0.21$     & $65.77 \pm 0.38$     & $75.98 \pm 0.31$     & $77.31 \pm 0.33$  \\ \hline
\end{tabular}
\caption{Results of 5-shot 5-way classification accuracy of our method on different sets of \di  using \cite{ravichandran2019few} as meta-learning algorithm. The evaluation is performed on 2000 episodes and we report the mean and the $95\%$ confidence interval.}

\label{Tab:mim55++}
}
\end{table*}

\renewcommand{\colwidthA}{2.2cm}
\begin{table*}[h]
\centering{
\small
\begin{tabular}{lccccc}
\hline
             & Seen classes from base domain     & Seen classes from incremental domain     &  Unseen classes from  base domain       & Unseen classes from incremental domain      & Unseen classes from both domains  \\ \hline
Base    & $56.32 \pm 0.50$    & $32.88 \pm 0.34$     & $37.90 \pm 0.37$     & $33.91 \pm 0.33$     & $39.56 \pm 0.40$     \\ \hline
Continued     & $37.12 \pm 0.36$    & $61.34 \pm 0.56$     & $32.81 \pm 0.32$     & $46.61 \pm 0.43$     & $42.91 \pm 0.42$     \\
Distillation      & $37.20 \pm 0.37$    & $60.23 \pm 0.55$     & $32.32 \pm 0.32$     & $46.12 \pm 0.44$     & $42.20 \pm 0.41$     \\
Identity   & $42.57 \pm 0.41$    & $59.31 \pm 0.55$     & $35.01 \pm 0.35$     & $45.55 \pm 0.43$     & $43.91 \pm 0.42$     \\
\hline
Full & $62.23 \pm 0.50$    & $67.25 \pm 0.53$     & $40.06 \pm 0.39$     & $50.78 \pm 0.45$     & $54.57 \pm 0.48$          \\ \hline
\end{tabular}
\caption{Results of \shot{1}{5} classification accuracy of our method on different sets of \di  using \cite{snell2017prototypical} as meta-learning algorithm. The evaluation is performed on 2000 episodes and we report the mean and the $95\%$ confidence interval. Base domain is natural and incremental domain is man-made classes.}
\label{Tab:mim51}
}
\end{table*}

\renewcommand{\colwidthA}{2.2cm}
\begin{table*}[h]
\centering{
\small
\begin{tabular}{lccccc}
\hline
             & Seen classes from base domain     & Seen classes from incremental Domain     &  Unseen classes from  base domain       & Unseen classes from  incremental domain      & Unseen classes from both domains  \\ \hline
Base   & $72.58 \pm 0.40$    & $39.68 \pm 0.37$     & $45.33 \pm 0.39$     & $40.43 \pm 0.34$     & $46.72 \pm 0.39$     \\ \hline
Continued     & $48.54 \pm 0.40$    & $76.46 \pm 0.43$     & $39.37 \pm 0.35$     & $54.93 \pm 0.41$     & $51.64 \pm 0.43$     \\
Distillation      & $49.83 \pm 0.40$    & $74.63 \pm 0.43$     & $40.36 \pm 0.36$     & $54.80 \pm 0.41$     & $52.32 \pm 0.43$     \\
Identity    & $69.82 \pm 0.41$    & $62.56 \pm 0.44$     & $46.22 \pm 0.39$     & $49.83 \pm 0.40$     & $53.03 \pm 0.41$     \\ \hline
Full & $70.10 \pm 0.42$    & $73.70 \pm 0.44$     & $46.81 \pm 0.40$     & $55.19 \pm 0.41$     & $57.97 \pm 0.42$          \\ \hline
\end{tabular}
\caption{Results of \shot{1}{5} classification accuracy of our method on different sets of \di  using \cite{ravichandran2019few} as meta-learning algorithm. The evaluation is performed on 2000 episodes and we report the mean and the $95\%$ confidence interval. Base domain is natural and incremental domain is man-made classes.}
\label{Tab:mim51++}
}
\end{table*}

\renewcommand{\colwidthA}{2.2cm}
\begin{table*}[h]
\centering{
\small
\begin{tabular}{lccccc}
\hline
             & Seen classes from base domain     & Seen classes from incremental domain     &  Unseen classes from  base domain       & Unseen classes from  incremental domain      & Unseen classes from both domains  \\ \hline
Base    & $91.75 \pm 0.18$    & $50.67 \pm 0.37$     & $70.94 \pm 0.35$     & $45.04 \pm 0.34$     & $62.57 \pm 0.44$     \\ \hline
Continued     & $71.41 \pm 0.31$    & $84.40 \pm 0.30$     & $57.11 \pm 0.32$     & $59.61 \pm 0.43$     & $63.40 \pm 0.38$     \\
Distillation      & $73.51 \pm 0.31$    & $83.31 \pm 0.31$     & $58.87 \pm 0.32$     & $59.64 \pm 0.42$     & $64.36 \pm 0.37$     \\
Identity   & $88.39 \pm 0.23$    & $81.24 \pm 0.34$     & $70.83 \pm 0.34$     & $60.82 \pm 0.39$     & $71.13 \pm 0.38$     \\
\hline
Full& $90.35 \pm 0.21$    & $87.11 \pm 0.27$     & $73.03 \pm 0.34$     & $62.44 \pm 0.42$     & $75.27 \pm 0.38$          \\ \hline
\end{tabular}
\caption{Results of \shot{5}{5} classification accuracy of our method on different sets of \di  using \cite{snell2017prototypical} as meta-learning algorithm. The evaluation is performed on 2000 episodes and we report the mean and the $95\%$ confidence interval. Base domain is man-made and incremental domain is natural classes.}
\label{Tab:mim552}
}
\end{table*}

\renewcommand{\colwidthA}{2.2cm}
\begin{table*}[h]
\centering{
\small
\begin{tabular}{lccccc}
\hline
             & Seen classes from base domain     & Seen classes from incremental domain     &  Unseen classes from  base domain       & Unseen classes from  incremental domain      & Unseen classes from both domains  \\ \hline
Base    & $72.45 \pm 0.51$    & $33.41 \pm 0.34$     & $51.48 \pm 0.48$     & $32.32 \pm 0.31$     & $43.88 \pm 0.45$     \\ \hline
Continued     & $46.08 \pm 0.42$    & $61.72 \pm 0.56$     & $39.90 \pm 0.39$     & $42.38 \pm 0.43$     & $44.85 \pm 0.43$     \\
Distillation      & $48.35 \pm 0.44$    & $58.13 \pm 0.56$     & $40.66 \pm 0.40$     & $41.30 \pm 0.41$     & $44.04 \pm 0.42$     \\
Identity & $58.98 \pm 0.47$    & $58.90 \pm 0.56$     & $46.33 \pm 0.44$     & $43.14 \pm 0.42$     & $48.33 \pm 0.44$     \\
\hline
Full & $73.31 \pm 0.50$    & $68.54 \pm 0.55$     & $52.29 \pm 0.45$     & $44.28 \pm 0.45$     & $58.11 \pm 0.50$          \\ \hline
\end{tabular}
\caption{Results of \shot{1}{5} classification accuracy of our method on different sets of \di  using \cite{snell2017prototypical} as meta-learning algorithm. The evaluation is performed on 2000 episodes and we report the mean and the $95\%$ confidence interval. Base domain is man-made and incremental domain is natural classes.}
\label{Tab:mim512}
}
\end{table*}

\renewcommand{\colwidthA}{2.2cm}
\begin{table*}[h]
\centering{
\small
\begin{tabular}{lccccc}
\hline
             & Seen classes from base domain     & Seen classes from incremental domain     &  Unseen classes from  base domain       & Unseen classes from  incremental domain      & Unseen classes from both domains  \\ \hline
Base    & $59.58 \pm 0.40$    & $91.91 \pm 0.16$     & $73.04 \pm 0.32$     & $53.20 \pm 0.37$     & $68.12 \pm 0.40$     \\ \hline
Continued     & $87.66 \pm 0.24$    & $72.11 \pm 0.31$     & $61.39 \pm 0.31$     & $63.87 \pm 0.40$     & $67.91 \pm 0.35$     \\
Distillation      & $82.45 \pm 0.31$    & $77.01 \pm 0.29$     & $62.91 \pm 0.32$     & $62.31 \pm 0.42$     & $67.79 \pm 0.36$     \\
Identity    & $79.80 \pm 0.33$    & $90.43 \pm 0.19$     & $74.04 \pm 0.31$     & $61.31 \pm 0.39$     & $72.60 \pm 0.37$     \\ \hline
Full & $85.66 \pm 0.28$    & $89.44 \pm 0.20$     & $74.60 \pm 0.32$     & $66.12 \pm 0.39$     & $76.92 \pm 0.35$          \\ \hline
\end{tabular}
\caption{Results of \shot{5}{5} classification accuracy of our method on different sets of \di  using \cite{ravichandran2019few} as meta-learning algorithm. The evaluation is performed on 2000 episodes and we report the mean and the $95\%$ confidence interval. Base domain is man-made and incremental domain is natural classes.}
\label{Tab:mim552++}
}
\end{table*}

\renewcommand{\colwidthA}{2.2cm}
\begin{table*}[h]
\centering{
\small
\begin{tabular}{lccccc}
\hline
             & Seen classes from base domain     & Seen classes form incremental domain     &  Unseen classes from  base domain       & Unseen classes from  incremental domain      & Unseen classes from both domains  \\ \hline
Base    & $42.27 \pm 0.38$    & $79.97 \pm 0.40$     & $53.49 \pm 0.42$     & $37.67 \pm 0.34$     & $49.74 \pm 0.42$ \\ \hline
Continued     & $73.37 \pm 0.47$    & $51.59 \pm 0.40$     & $42.22 \pm 0.35$     & $47.09 \pm 0.42$     & $48.64 \pm 0.41$     \\
Distillation      & $66.49 \pm 0.50$    & $56.71 \pm 0.42$     & $43.68 \pm 0.36$     & $46.29 \pm 0.43$     & $49.12 \pm 0.41$     \\
Identity    & $60.64 \pm 0.46$    & $76.91 \pm 0.41$     & $54.02 \pm 0.42$     & $44.01 \pm 0.39$     & $53.16 \pm 0.42$     \\ \hline
Full & $68.86 \pm 0.47$    & $73.87 \pm 0.43$     & $53.91 \pm 0.42$     & $48.00 \pm 0.42$     & $58.00 \pm 0.43$          \\ \hline
\end{tabular}
\caption{Results of \shot{1}{5} classification accuracy of our method on different sets of \di  using \cite{ravichandran2019few} as meta-learning algorithm. The evaluation is performed on 2000 episodes and we report the mean and the $95\%$ confidence interval. Base domain is man-made and incremental domain is natural classes.}
\label{Tab:mim512++}
}
\end{table*}

\newpage

\appendix

\subsection*{When we store the old data}

We can then write the objective of incremental meta-learning as:
\begin{align}
L(w_{t+1}) &= \underbrace{\sum_{\T \in \M_\new} \ell(w_{t+1}, \T)}_{\text{Classification accuracy on new tasks}} \nonumber\\
&\quad\quad\quad+ \lambda \underbrace{\sum_{\T \in \M_\text{base}} d_q\big( p^\wtp_{\T}(y|x), p^\wt_{\T}(y|x) \big)}_{\text{Distance of classifiers of base tasks computed on base and new examples}}\nonumber\\
&= L_{\M_\new}(w_{t+1}) + \lambda \big( L_{d_{q_\text{base}},
 \M_\text{base}}(w_{t+1}, w_t) +  L_{d_{q_\new}, \M_\text{base}}(w_{t+1}, w_t) \big) \label{incremental-loss}
\end{align}
Where the first term is the usual meta-training loss on the new tasks $\M_\new$, and the second two terms are regularizer that forces the new meta-algorithm at time $t+1$ to output similar classifiers as the one at time $t$. This corresponds to the loss just above Section 5 in the arXiv version of the paper.

\subsection*{When we throw away base data}

In this case, we store the classifiers on the base class (since we can not compute them from scratch every time, not having access to the base data), and we take $q(x) = q_\new(x)$ to just sample new images in order to probe the classifier (again since we do not have the old data).
Using the (not very happy) notation $p^\wt_{\T}(y|f_\wtp(x))$ to mean running the old discriminant with the \textit{old} decision boundary (which in protonets means the one computed with the precomputed centers at time $t$) but embedding the data with the new embedding $f_\wtp(x)$. This translates in requiring that the new embedding does not move too much with respect to the old decision boundaries, but gives the embeddinng freedom to move in their nullspace.
\begin{align}
L(w_{t+1}) &= \sum_{\T \in \M_\new} \ell(w_{t+1}, \T) + \lambda \sum_{\T \in \M_\text{base}} d_{q_\new}\big( p^\wt_{\T}(y|f_\wtp(x)), p^\wt_{\T}(y|f_\wt(x)) \big)  \label{eq:throw-loss}
\end{align}
where again $p^\wt_{\T}(y|f_\wtp(x))$ means the classifier with stored boundary (equivalent, centers) at time $t$ evaluated on data embedded with the new embedding $f_\wtp(x)$.

\renewcommand{\colwidthA}{1.8cm}
\begin{table*}[t!]
\caption{Comparison of the two proposed methods across multiple datasets.}
\centering{
\small
\resizebox{0.95\textwidth}{!}{
\begin{tabular}{l|l|lll|lll}
\hline
\multirow{2}{*}[-0.5em]{\ds} &  \multirow{2}{*}{Method} & \multicolumn{3}{c|}{\shot{1}{5}} & \multicolumn{3}{c}{\shot{5}{5}}   \\
\cline{3-8}
                       & & \multicolumn{1}{C{\colwidthA}}{Old classes (32)} & \multicolumn{1}{C{\colwidthA}}{New classes (32) } & \multicolumn{1}{C{\colwidthA}|}{Unseen classes (20)} & \multicolumn{1}{C{\colwidthA}}{Old classes (32)} & \multicolumn{1}{C{\colwidthA}}{New classes (32)} & \multicolumn{1}{C{\colwidthA}}{Unseen classes (20)}\\ 
                       \hline 
\multirow{2}{*}{\mi} 
& EIML & $\mathbf{68.95 \pm 0.50}$   & $71.43 \pm 0.52$     & $54.86 \pm 0.42$ & $\mathbf{90.20 \pm 0.20}$  & $86.91 \pm 0.25$ & $74.39 \pm 0.50$ \\
& IFA    & $66.54 \pm 0.49$    & $\mathbf{71.92 \pm 0.51}$     & $\mathbf{55.22 \pm 0.43}$ & $89.14 \pm 0.21$  & $\mathbf{87.32 \pm 0.25}$ & $\mathbf{75.11 \pm 0.31}$ \\
\hline
\multirow{2}{*}{\ti}  
& EIML & $72.50 \pm 0.51$    & $69.44 \pm 0.52$     & $58.42 \pm 0.50$ & $88.93 \pm 0.27$ & $86.25 \pm 0.32$ & $77.97 \pm 0.42$ \\
& IFA    & $\mathbf{72.65 \pm 0.51}$    & $\mathbf{70.17 \pm 0.53}$     & $\mathbf{58.71 \pm 0.50}$ & $\mathbf{89.03 \pm 0.27}$  & $\mathbf{86.91 \pm 0.31}$ & $\mathbf{78.40 \pm 0.42}$     \\ 
\hline
\multirow{2}{*}{\di} 
& EIML & $\mathbf{48.48 \pm 0.44}$ & $\mathbf{42.27 \pm 0.41}$ &$ 43.91 \pm 0.42$ &  $\mathbf{83.23 \pm 0.25}$ & $\mathbf{65.81 \pm 0.40}$ & $69.99\pm 0.36$  \\
& IFA & $42.57 \pm 0.41$ & $38.95 \pm 0.39$ & $\mathbf{44.88 \pm 0.43}$ & $81.26 \pm 0.27$ & $65.78 \pm 0.4$1 & $\mathbf{70.36\pm 0.36}$\\
\hline
\end{tabular}}
\label{Tab:comaprison}
}
\end{table*}

\begin{table*}[t!]
\caption{Classification accuracy on different sets of \ti using IML and meta-training using Embedded Class Models \cite{ravichandran2019few}. We show this performance on three different sets: tasks sampled from the base classes, from  incremental classes and tasks sampled from the meta-test sets.}
\centering{
\small
\resizebox{0.95\textwidth}{!}{
\begin{tabular}{l|lll|lll|l}
\hline
\multirow{2}{*}{Model} & \multicolumn{3}{c|}{\shot{1}{5}} & \multicolumn{3}{c|}{\shot{5}{5}} & \multirow{2}{*}{Old data} \\
\cline{2-7}
                       & \multicolumn{1}{C{\colwidthA}}{Old classes (176)} & \multicolumn{1}{C{\colwidthA}}{New classes (175) } & \multicolumn{1}{C{\colwidthA}|}{Unseen classes (160)} & \multicolumn{1}{C{\colwidthA}}{Old classes (176)} & \multicolumn{1}{C{\colwidthA}}{New classes (175) } & \multicolumn{1}{C{\colwidthA}|}{Unseen classes (160) } &                \\
                       \hline
\rowcolor{orange!20}  Base   & $75.79 \pm 0.43$   &  $67.28 \pm 0.47$   & $62.66 \pm 0.46$  & $89.75 \pm 0.24$   & $83.96 \pm 0.33$  & $80.90 \pm 0.35$  & $-$                \\ \hline
Baseline    & &     &  &   & &  & $0$                \\
DFA      &    &     &  &  &  &  & $0$                \\
IFA    &     &      &  &   &  &  &  $1/600$            \\ \hline
\rowcolor{green!20}  Paragon & $75.86 \pm 0.42$     & $74.99 \pm 0.45$     & $65.78 \pm 0.46$ & $89.82 \pm 0.25$  & $89.17 \pm 0.27$  & $83.33 \pm 0.33$  &  $600/600$          \\ \hline\end{tabular}}
\label{Tab:tiproto++}
}
\end{table*}